\PassOptionsToPackage{unicode}{hyperref}
\PassOptionsToPackage{hyphens}{url}
\PassOptionsToPackage{dvipsnames,svgnames,x11names}{xcolor}
\documentclass[
]{article}
\usepackage{xcolor}
\usepackage{amsmath,amssymb}
\usepackage{iftex}
\ifPDFTeX
  \usepackage[T1]{fontenc}
  \usepackage[utf8]{inputenc}
  \usepackage{textcomp} 
\else 
  \usepackage{unicode-math} 
  \defaultfontfeatures{Scale=MatchLowercase}
  \defaultfontfeatures[\rmfamily]{Ligatures=TeX,Scale=1}
\fi
\usepackage{lmodern}
\ifPDFTeX\else
\fi
\IfFileExists{upquote.sty}{\usepackage{upquote}}{}
\IfFileExists{microtype.sty}{
  \usepackage[]{microtype}
  \UseMicrotypeSet[protrusion]{basicmath} 
}{}
\makeatletter
\@ifundefined{KOMAClassName}{
  \IfFileExists{parskip.sty}{%
    \usepackage{parskip}
  }{
    \setlength{\parindent}{0pt}
    \setlength{\parskip}{6pt plus 2pt minus 1pt}}
}{
  \KOMAoptions{parskip=half}}
\makeatother
\makeatletter
\ifx\paragraph\undefined\else
  \let\oldparagraph\paragraph
  \renewcommand{\paragraph}{
    \@ifstar
      \xxxParagraphStar
      \xxxParagraphNoStar
  }
  \newcommand{\xxxParagraphStar}[1]{\oldparagraph*{#1}\mbox{}}
  \newcommand{\xxxParagraphNoStar}[1]{\oldparagraph{#1}\mbox{}}
\fi
\ifx\subparagraph\undefined\else
  \let\oldsubparagraph\subparagraph
  \renewcommand{\subparagraph}{
    \@ifstar
      \xxxSubParagraphStar
      \xxxSubParagraphNoStar
  }
  \newcommand{\xxxSubParagraphStar}[1]{\oldsubparagraph*{#1}\mbox{}}
  \newcommand{\xxxSubParagraphNoStar}[1]{\oldsubparagraph{#1}\mbox{}}
\fi
\makeatother

\usepackage{longtable,booktabs,array}
\usepackage{calc} 
\usepackage{etoolbox}
\makeatletter
\patchcmd\longtable{\par}{\if@noskipsec\mbox{}\fi\par}{}{}
\makeatother
\IfFileExists{footnotehyper.sty}{\usepackage{footnotehyper}}{\usepackage{footnote}}
\makesavenoteenv{longtable}
\usepackage{graphicx}
\makeatletter
\newsavebox\pandoc@box
\newcommand*\pandocbounded[1]{
  \sbox\pandoc@box{#1}%
  \Gscale@div\@tempa{\textheight}{\dimexpr\ht\pandoc@box+\dp\pandoc@box\relax}%
  \Gscale@div\@tempb{\linewidth}{\wd\pandoc@box}%
  \ifdim\@tempb\p@<\@tempa\p@\let\@tempa\@tempb\fi
  \ifdim\@tempa\p@<\p@\scalebox{\@tempa}{\usebox\pandoc@box}%
  \else\usebox{\pandoc@box}%
  \fi%
}
\def\fps@figure{htbp}
\makeatother

\providecommand{\tightlist}{%
  \setlength{\itemsep}{0pt}\setlength{\parskip}{0pt}}

\usepackage[]{biblatex}
\usepackage{booktabs}
\usepackage{caption}
\usepackage{longtable}
\usepackage{colortbl}
\usepackage{array}
\usepackage{anyfontsize}
\usepackage{multirow}
\usepackage{arxiv}
\usepackage{orcidlink}
\usepackage{amsmath}
\usepackage[T1]{fontenc}
\makeatletter
\@ifpackageloaded{caption}{}{\usepackage{caption}}
\AtBeginDocument{%
\ifdefined\contentsname
  \renewcommand*\contentsname{Table of contents}
\else
  \newcommand\contentsname{Table of contents}
\fi
\ifdefined\listfigurename
  \renewcommand*\listfigurename{List of Figures}
\else
  \newcommand\listfigurename{List of Figures}
\fi
\ifdefined\listtablename
  \renewcommand*\listtablename{List of Tables}
\else
  \newcommand\listtablename{List of Tables}
\fi
\ifdefined\figurename
  \renewcommand*\figurename{Figure}
\else
  \newcommand\figurename{Figure}
\fi
\ifdefined\tablename
  \renewcommand*\tablename{Table}
\else
  \newcommand\tablename{Table}
\fi
}
\@ifpackageloaded{float}{}{\usepackage{float}}
\floatstyle{ruled}
\@ifundefined{c@chapter}{\newfloat{codelisting}{h}{lop}}{\newfloat{codelisting}{h}{lop}[chapter]}
\floatname{codelisting}{Listing}

\makeatother
\makeatletter
\@ifpackageloaded{caption}{}{\usepackage{caption}}
\@ifpackageloaded{subcaption}{}{\usepackage{subcaption}}
\makeatother
\usepackage{bookmark}
\IfFileExists{xurl.sty}{\usepackage{xurl}}{} 
\hypersetup{
  pdftitle={Trust in Autonomous Human-Robot Collaboration: Effects of Responsive Interaction Policies},
  pdfauthor={Shauna Heron; Meng Cheng Lau},
  pdfkeywords={Human-robot collaboration, Human-robot
interaction, socially assistive robotics, autonomous robot
systems, embodied AI, trust in human-robot interaction, affect-adaptive
robot systems},
  colorlinks=true,
  linkcolor={blue},
  filecolor={Maroon},
  citecolor={Blue},
  urlcolor={Blue},
  pdfcreator={LaTeX via pandoc}}

\newcommand{\runninghead}{A Preprint }
\renewcommand{\runninghead}{A PREPRINT }
\title{Trust in Autonomous Human-Robot Collaboration: Effects of
Responsive Interaction Policies}
\usepackage{etoolbox}
\makeatletter
\providecommand{\subtitle}[1]{
  \apptocmd{\@title}{\par {\large #1 \par}}{}{}
}
\makeatother
\subtitle{An In-Person Pilot Study}
\def\asep{\\\\\\ } 
\author{\textbf{Shauna Heron}~\orcidlink{0000-0002-9262-6718}\\School of
Social Sciences\\Laurentian University\\Sudbury, ON,\ P3E
2C6\\\href{mailto:sheron@laurentian.ca}{sheron@laurentian.ca}\asep\textbf{Meng
Cheng Lau}~\orcidlink{0000-0003-3517-4900}\\School of Engineering and
Computer Science\\Laurentian University\\Sudbury, ON,\ P3E
2C6\\\href{mailto:mclau@laurentian.ca}{mclau@laurentian.ca}}
\date{}
\begin{document}
\maketitle
\begin{abstract}
Trust plays a central role in human-robot collaboration, yet its
formation is rarely examined under the constraints of fully autonomous
interaction. This pilot study investigated how interaction policy
influences trust during in-person collaboration with a social robot
operating without Wizard-of-Oz control or scripted repair. Participants
completed a multi-stage collaborative task with a mobile robot that
autonomously managed spoken-language dialogue, affect inference, and
task progression. Two interaction policies were compared: a responsive
policy, in which the robot proactively adapted its dialogue and
assistance based on inferred interaction state, and a neutral, reactive
policy, in which the robot provided only direct, task-relevant responses
when prompted. Responsive interaction was associated with significantly
higher post-interaction trust under viable communication conditions,
despite no reliable differences in overall task accuracy. Sensitivity
analyses indicated that affective and experiential components of trust
were more sensitive to communication breakdown than evaluative judgments
of reliability, and that as language-mediated interaction degraded, the
trust advantage associated with responsiveness attenuated and ratings
became less clearly interpretable as calibrated evaluations of
collaborative competence. These findings suggest that trust in
autonomous human-robot interaction emerges from process-level
interaction dynamics and operates within constraints imposed by
communication viability, highlighting the importance of evaluating trust
under real autonomy conditions when designing interactive robotic
systems.
\end{abstract}
{\bfseries \emph Keywords}
\def\sep{\textbullet\ }
Human-robot collaboration \sep Human-robot interaction \sep socially
assistive robotics \sep autonomous robot systems \sep embodied
AI \sep trust in human-robot interaction \sep 
affect-adaptive robot systems

\section{Introduction}\label{introduction}

As artificial intelligence (AI) technologies advance, they are
increasingly integrated into robotic systems, enabling more adaptive,
autonomous, and context-sensitive behaviour in real-world environments.
This convergence has accelerated the deployment of robots across
safety-critical domains such as manufacturing \autocite{choi2025},
mining \autocite{fu2021}, and healthcare \autocite{ciuffreda2025}, where
robots are now expected to operate alongside humans rather than in
isolation \autocite{diab2025,spitale2023}. In these collaborative
settings, successful deployment depends not only on technical
performance and safety guarantees, but also on whether human users are
willing to rely on, communicate with, and coordinate their actions
around robotic partners \autocite{campagna2025,emaminejad2022}. Trust
has therefore emerged as a central determinant of adoption and effective
use in human-robot collaboration (HRC)
\autocite{wischnewski2023,campagna2025}. Insufficient trust can lead to
disuse or rejection of automation, while excessive trust risks
overreliance and accidents---particularly in environments characterized
by uncertainty or incomplete information \autocite{devisser2020}.

In HRC, trust is commonly understood as a willingness to rely on an
agent under conditions of uncertainty and risk
\autocite{muir1994,hancock2011}. This reliance is dynamically
calibrated, shaping how closely users monitor a robot, when they
intervene, and whether they defer or override its actions. Appropriately
calibrated trust supports effective coordination, whereas under-trust
may result in disengagement or redundant oversight, and over-trust can
lead to inappropriate reliance and unsafe outcomes
\autocite{devisser2020}. These dynamics are especially pronounced in
dialogue-driven collaborative tasks, where misunderstandings, delays, or
ambiguous responses may directly influence users' ongoing assessments of
a robot's competence and reliability.

A substantial body of human-robot interaction (HRI) research has
examined how robot behaviour influences user trust, perceived
reliability, and cooperation across industrial and social contexts
\autocite{shayganfar2019,fartook2025}. Trust is typically conceptualized
as a multidimensional construct encompassing cognitive evaluations of
competence, predictability, and reliability, alongside a behavioural
willingness to collaborate toward shared goals under conditions of risk
or uncertainty \autocite{muir1994,hancock2011,devisser2020}. Despite
this multidimensional framing, empirical studies have most often
operationalized trust using post-interaction self-report questionnaires
collected following short, highly controlled, and often scripted
interactions. While such measures provide valuable global assessments of
user attitudes, they offer limited insight into how trust is negotiated,
disrupted, and repaired during ongoing interaction---particularly in
autonomous systems where errors and ambiguities are unavoidable
\autocite{maure}.

Studying trust as an interactional process therefore requires
experimental settings in which users engage with robots that exhibit
both adaptive behaviour and realistic system limitations
\autocite{campagna2025}. In such settings, trust is shaped not only by
task success but by how robots handle uncertainty, errors, and
misalignment during interaction. Fully autonomous systems, where
dialogue management and response generation occur without human
intervention, provide a critical testbed for examining these dynamics,
as they expose users to the same constraints and breakdowns encountered
in real-world deployment \autocite{campagna2025}.

Yet in practice, much of the existing HRI trust literature has relied on
scripted behaviours, simulated environments, or Wizard-of-Oz paradigms
in which a human operator covertly manages the robot's behaviour
\autocite{bettencourt2025,campagna2025}. While these approaches are
valuable for isolating specific design factors, they obscure the
interaction breakdowns and system imperfections that characterize
deployed autonomous robots. Limitations such as speech recognition
errors, delayed responses, misinterpretations of user intent, and
incomplete affect sensing are not peripheral issues but defining
features of real-world interaction. These failures are likely to play a
decisive role in shaping trust and collaboration, yet remain
underrepresented in empirical evaluations \autocite{campagna2025}.

Within HRI, a range of design strategies have been proposed to support
appropriate trust calibration during collaboration, including robot
appearance \autocite{nicolas2021}, transparency cues
\autocite{zhang2019}, explanations of system intent, adaptive feedback
\autocite{val-calvo2020}, and interaction pacing \autocite{kok2020}.
Many of these approaches aim to help users form accurate expectations
about a robot's capabilities and limitations, particularly in contexts
involving uncertainty or partial observability. Among these strategies,
adaptive interaction behaviour---how and when a robot responds to user
state and task context---has been identified as a particularly
influential factor in shaping perceptions of competence, reliability,
and collaboration \autocite{fartook2025}.

Recent advances in AI have expanded the range of interaction strategies
available to autonomous robotic platforms in practice, enabling systems
to move beyond fixed, scripted behaviours toward adaptive interaction
policies that respond to user state and task context in real time
\autocite{atone2022}. Improvements in spoken-language processing,
dialogue state tracking, and large language model-based reasoning now
allow robots to adjust not only what they say, but when and how
assistance is provided during collaboration \autocite{wei}. In parallel,
advances in affect inference from language and interaction cues have
made it increasingly feasible for robots to incorporate estimates of
user emotional state into interaction management
\autocite{mcduff,spitale2023}. As a result, responsiveness in
contemporary HRI is increasingly understood as a property of an
underlying interaction policy, governing how a robot interprets cues,
initiates support, and manages uncertainty, rather than as a
surface-level social behaviour
\autocite{birnbaum2016,shayganfar2019,fartook2025}.

From an engineering perspective, responsiveness in autonomous
human-robot interaction is not implemented as a single behavioural rule
or surface-level cue, but emerges from the coordination of multiple
system components responsible for perception, interpretation, and action
\autocite{arkin2003}. In practice, responsive behaviour depends on the
integration of dialogue management, state tracking, and inference
mechanisms that estimate task progress, interaction uncertainty, and
user affect. Proactive assistance based on this integrated state
representation differs fundamentally from reactive, request-based
behaviour; for example, a robot may offer clarification, encouragement,
or pacing adjustments when confusion, hesitation, or frustration is
inferred, rather than waiting for an explicit request for help
\autocite{birnbaum2016}.

Implementing such behaviour requires autonomous systems to manage
spoken-language dialogue, maintain interaction state over time, and
coordinate verbal and nonverbal responses in real time, all while
operating under noise, latency, and sensing uncertainty
\autocite{campagna2025}. As a result, responsiveness in deployed systems
reflects properties of the overall interaction architecture, not merely
the presence or absence of adaptive dialogue. Despite growing interest
in responsive and affect-aware robots, relatively little empirical work
has examined how such integrated interaction systems operate under fully
autonomous conditions, or how their behaviour shapes trust and
collaboration when communication breakdowns cannot be externally
repaired \autocite{campagna2025}.

The present pilot study addresses these gaps by examining trust and
collaboration during in-person interaction with a fully autonomous
social robot in which participants collaborated with one of two versions
of the same robot platform during a dialogue-driven puzzle task
requiring shared problem solving. In both conditions, all interaction
management, including speech recognition, dialogue state tracking, task
progression, and response generation was handled and logged autonomously
by a centralized dialogue agent without human intervention. In the
responsive condition, the robot employed a proactive, affect-aware
interaction policy, adapting its assistance based on conversational cues
and inferred user affect (e.g., frustration or engagement). In the
neutral condition, the robot followed a reactive policy, providing basic
guidance and periodic check-ins, otherwise providing assistance only
when explicitly requested.

The pilot study had three primary objectives: (1) to design and evaluate
the feasibility of an autonomous spoken-language interaction system with
affect-responsive behaviour on a mobile robot platform; (2) to assess
whether a responsive interaction policy influences post-interaction
trust and collaborative experience under autonomous conditions; and (3)
to explore how behavioural and interaction-level indicators align with
subjective trust evaluations.

Rather than optimizing for flawless interaction, this pilot study
prioritizes feasibility assessment and mechanism exploration over
confirmatory hypothesis testing. The sample size is modest, interaction
conditions subject to real-world autonomy constraints, and final
condition assignment was not fully randomized due to participant
attrition. Accordingly, results should be interpreted as exploratory
evidence regarding how interaction policy shapes trust under autonomous
spoken-language collaboration, and as guidance for the design of larger,
confirmatory studies. The findings are intended to inform the design of
a larger subsequent study by evaluating feasibility and identifying
technical, interactional, and methodological challenges that must be
addressed when evaluating affect-responsive robots in real-world
contexts.

\section{Methods}\label{methods}

This study employed a between-subjects experimental design to examine
how robot interaction policy influences trust and collaboration during
fully autonomous, in-person human-robot interaction. The sole
experimental factor was the robot's interaction policy. Participants
were initially randomly assigned to condition; however, final condition
allocation was partially constrained by participant attrition and
scheduling, as described below.

Throughout this paper, references to ``the robot'' denote the autonomous
interactive system comprising the Misty-II hardware platform and an
integrated offboard software pipeline. Spoken-language understanding,
dialogue management, task logic, and interaction policy execution were
handled on an external edge device which interfaced with the robot via
websocket interfaces. The Misty-II platform was responsible for audio
capture, text to speech, and the execution of embodied behaviours
including facial expressions, body movement, and LED signalling. Despite
this distributed execution, all interaction decisions were generated
autonomously by an AI driven centralized dialogue agent responsible for
coordinating spoken-language understanding, task state, and verbal and
nonverbal behaviour based on the interaction policy. More detail on the
robot platform and software architecture is provided in Appendix A.

\subsection{Study Protocol}\label{study-protocol}

Participants signed up for the study, provided informed consent, were
randomly assigned to a group, and completed a pre-session questionnaire
before their in-person session via a survey hosted on Qualtrics. The
elapsed time between sign-up and the in-person session was at most a
week and at minimum immediately before the session. The pre-session
questionnaire collected basic demographics information and assessed
baseline characteristics, including the Negative Attitudes Toward Robots
Scale (NARS) and the short form of the Need for Cognition scale (NFC-s)
\autocite{nomura,cacioppo1982}. These measures were used to capture a
baseline of individual differences that may moderate responses to robot
interaction.

In-person sessions were conducted in a quiet, private room at Laurentian
University between November and December 2025. Prior to each session,
the robot's interaction policy was configured to the assigned
experimental condition.

Upon arrival, participants were greeted by the researcher, provided with
a brief overview of the session, and given instructions for effective
communication with the robot. Once participants indicated readiness, the
researcher exited the room, leaving the participant and robot to
complete the interaction without human presence or observation (see
Figure~\ref{fig-task1}). Participants initiated the interaction by
clicking a start button on the interface and were informed that they
could terminate the session at any time without penalty.

\begin{figure}

\centering{

\includegraphics[width=4.375in,height=\textheight,keepaspectratio]{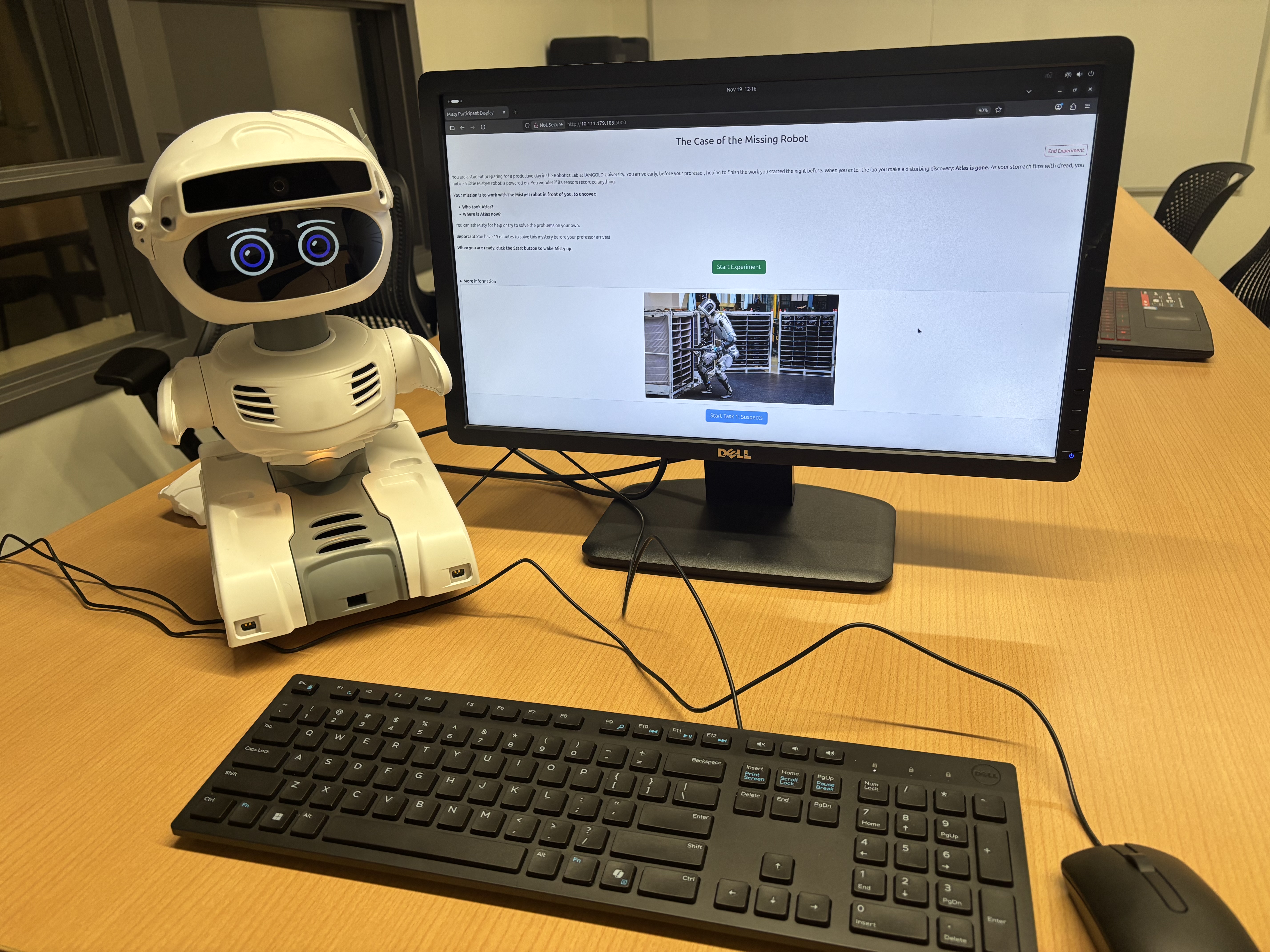}

}

\caption{\label{fig-setup}Experimental setup showing the autonomous
robot and participant-facing task interface used during in-person
sessions. Participants entered task responses and navigated between task
stages using the interface, while the robot autonomously tracked task
state and adapted its interaction based on participant input.}

\end{figure}%

Following task completion, participants completed a 21-item
post-interaction questionnaire assessing trust on a provied laptop.
Participants then engaged in a brief debrief with the researcher and
were awarded a \$15 gift card. Total session duration averaged
approximately 30 minutes.

\subsection{Interaction Policies}\label{interaction-policies}

Across both tasks, interaction behaviour was governed by one of two
interaction policies that differed in how the robot was intended to
initiate, frame, and adapt its contributions during collaboration. Under
both policies, the robot continuously monitored interaction timing and
issued brief check-ins following extended periods of participant silence
in order to preserve interaction continuity. In the neutral, reactive
condition, these check-ins were minimal and task-focused, serving only
to signal availability without providing guidance, encouragement, or
additional framing (e.g., ``I'm ready for your next question.'').

In contrast, under the responsive policy, the robot's utterances were
affectively framed and context-sensitive. In addition to answering
questions, the robot proactively adapted its dialogue, assistance, and
repair strategies based on inferred interaction state, such as
hesitation, frustration, or apparent difficulty. This included
acknowledging task difficulty, offering encouragement, and proposing
collaborative reasoning (e.g., ``I can tell you're frustrated, don't
worry! we can reason through this together''), rather than waiting for
an explicit request for help. Beyond differences in check-in style, the
responsive robot also initiated guidance or clarification when
interaction stagnated, whereas the neutral robot limited its
contributions to basic task guidance and direct queries.

Both conditions operated fully autonomously without human intervention
and otherwise used the same underlying task logic and sensing
infrastructure.

\subsection{Collaborative Task Design}\label{collaborative-task-design}

The task structure was modelled after the interactive and immersive
puzzle game design by \textcite{lin2022} to examine collaboration under
two distinct dependency conditions: enforced collaboration, in which
successful task completion required the robot's involvement, and
optional collaboration, in which participants could choose whether and
how to engage the robot. To operationalize these conditions in a
naturalistic way, we embedded both tasks within a single immersive,
narrative-driven puzzle scenario.

Participants were positioned as investigators searching for a missing
laboratory robot named Atlas. The Misty-II robot served as a diegetic
guide and collaborative partner throughout the session, framing each
stage of the interaction as part of the unfolding investigation. This
narrative structure provided continuity across tasks while allowing the
dependency manipulation to vary between stages without disrupting
immersion. The full session lasted approximately 25 minutes and
consisted of five sequential stages, including two timed reasoning
tasks.

Interactions took place in a shared physical workspace that included the
Misty-II social robot and a participant-facing computer interface
\autocite{mistyrobotics} at Laurentian University. The interface
displayed task materials, recorded participant responses, and supported
progression through the game. Importantly, it did not function as a
control mechanism for the robot. Instead, the robot autonomously
monitored task progression and participant inputs via the interface,
adapting its dialogue and behaviour accordingly.

The interaction began with a brief greeting phase during which the robot
introduced itself and engaged in rapport-building dialogue (e.g., ``Hi,
I'm Misty. What's your name?''). This was followed by a mission briefing
outlining the investigative scenario and establishing the collaborative
goal: to determine who had taken Atlas and where it might now be
located.

Participants then completed two core task stages. The first was a
robot-dependent collaborative reasoning task in which Misty's
participation was required to solve the problem. The second was a more
open-ended and complex problem-solving task in which robot assistance
was available but not required. The interaction concluded with a wrap-up
stage in which the robot provided closing feedback and formally ended
the session.

Participants advanced between stages using the interface, either in
response to the robot's prompts or at their own discretion. All spoken
dialogue and interaction events were managed autonomously by the robot
and logged automatically for analysis.

\subsubsection{Task 1: Robot-Dependent Collaborative
Reasoning}\label{task-1-robot-dependent-collaborative-reasoning}

In the first task, participants were asked to identify who had taken the
missing robot from a 6 × 4 grid of 24 potential ``suspects.'' Within the
narrative, Misty informed participants that she had access to internal
security footage but was unable to directly reveal the suspect's
identity due to laboratory security protocols. Instead, she could answer
only yes/no questions about observable features of the individual in the
footage (e.g., ``Was the suspect wearing a hat?''). This constraint
established the robot as the sole source of task-relevant information
while maintaining narrative plausibility. The suspect grid was displayed
on the interface and allowed participants to click on and grey out
eliminated candidates, while all questions were posed verbally to the
robot (see Figure~\ref{fig-task1}).

The robot possessed the ground-truth information necessary to answer
each question correctly. Successful task completion was therefore fully
dependent on interaction with the robot, creating an enforced
collaborative dynamic. Participants were required to coordinate a
questioning strategy that progressively eliminated candidates based on
shared features, narrowing the search space within a five-minute time
limit. Efficient performance depended on selecting informative questions
(i.e., features that divided the remaining candidates), tracking
eliminations, and adapting subsequent questions based on prior answers.

\begin{figure}

\centering{

\includegraphics[width=4.375in,height=\textheight,keepaspectratio]{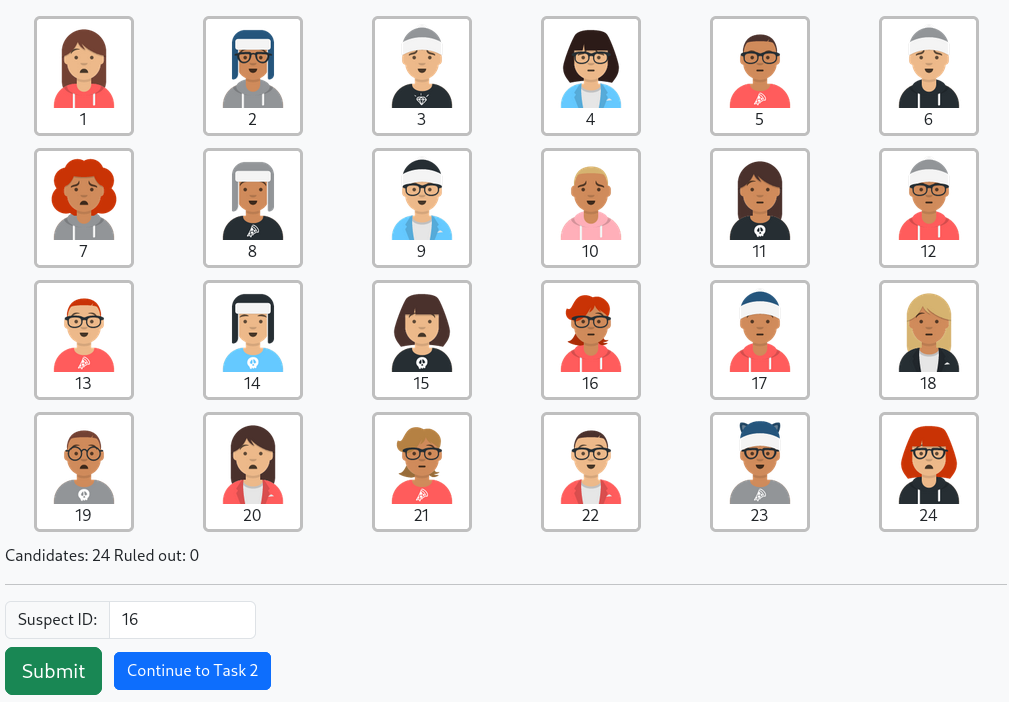}

}

\caption{\label{fig-task1}Task 1 interface including the 6 \(×\) 4 grid
of 24 candidates. Participants could track those eliminated by clicking
on subjects which would grey them out. A box was provided to input their
final answer and a button included to move to the next task.}

\end{figure}%

This structure made the task sensitive to interaction quality.
Inefficient questioning, repeated queries, or uncertainty about next
steps could slow progress and increase cognitive load, whereas effective
coordination supported rapid elimination and convergence on a solution.
The constrained yes/no format ensured consistent informational bandwidth
across participants and conditions, while still allowing meaningful
variation in collaboration strategy. The robot's behaviour under the
responsive policy was designed to support this process by providing
adaptive guidance, encouragement, and clarification when participants
appeared to struggle or become disengaged, whereas the neutral robot
provided only direct answers without additional support.

\subsubsection{Task 2: Open-Ended Collaborative Problem
Solving}\label{task-2-open-ended-collaborative-problem-solving}

The second task involved a more open-ended reasoning challenge designed
to shift the dependency structure from enforced to optional
collaboration. Participants were presented with multiple technical logs
through a simulated terminal interface that could be used to infer the
location of the missing robot. The task was intentionally cryptic and
difficult to solve within the allotted ten minutes without synthesizing
information across multiple sources.

The logs contained partial and indirect clues related to the robot's
activity, such as wireless connectivity records, sensor readings, and
timestamped system events. Solving the task required participants to
identify which logs were relevant, extract spatial or temporal cues, and
integrate these signals to progressively narrow down plausible
locations. As in Task 1, successful performance depended on managing
uncertainty and iteratively refining hypotheses rather than on
recognizing a single explicit cue.

Critically, unlike Task 1, the robot did not possess ground-truth
information about the solution or direct access to the contents of the
logs. This constraint was intentional: Misty's assistance was limited to
general reasoning support derived from its language model, such as
explaining how to interpret log formats, suggesting strategies for
cross-referencing timestamps, or prompting participants to reconcile
inconsistencies across sources. Participants could complete the task
independently or solicit assistance from the robot at their discretion
\autocite{lin2022}, allowing collaboration to emerge voluntarily rather
than being structurally enforced.

\subsection{Measures}\label{measures}

A combination of self-report and objective measures was used to assess
trust, engagement, and task performance.

\subsubsection{Self-Report Measures}\label{self-report-measures}

Participants completed a pre-session questionnaire assessing baseline
characteristics, including the Negative Attitudes Toward Robots Scale
(NARS) and the short form of the Need for Cognition scale (NFC-s)
\autocite{cacioppo1982,nomura}. These measures were used to capture a
baseline of individual differences that may moderate responses to robot
interaction.

Trust was assessed using two established self-report instruments
commonly used in human-robot interaction research: the Trust Perception
Scale-HRI (TPS-HRI) and the Trust in Industrial Human-Robot
Collaboration scale (TI-HRC) \autocite{schaefer2016,charalambous2016}.
Both measures were adapted to reflect the specific dialogue-driven task
context and interaction modality of the present study, while preserving
the original constructs and response intent of each scale. 9 items were
retained from the TI-HRC and 12 items from the TPS-HRI. Item wording was
modified to reference the robot's behaviour during the dialogue-driven
collaborative tasks and response formats were adjusted to ensure
interpretability for participants without prior robotics experience (see
Appendix B for a full item list). Internal consistency of the adapted
trust measures was evaluated within the present sample to assess
measurement reliability under the modified wording and response formats.
Given the pilot sample size, reliability estimates should be interpreted
cautiously, but are reported to provide transparency regarding scale
performance under fully autonomous interaction conditions.

All self-report scales demonstrated acceptable to strong internal
consistency in the present sample. The Negative Attitudes Toward Robots
scale (NARS) showed good reliability (\(\alpha\) = .81), with subscale
alphas ranging from .72 to .76. The Need for Cognition scale (NFC)
demonstrated strong internal consistency (\(\alpha\) = .87). Both
post-interaction trust measures also exhibited high reliability. The
Trust Perception Scale-HRI (TPS-HRI) showed strong internal consistency
overall (\(\alpha\) = .89), with subscale reliabilities ranging from .74
to .88. The Trust in Industrial Human-Robot Collaboration scale (TI-HRC)
demonstrated very high internal consistency (\(\alpha\) = .95).

Together, these instruments capture complementary dimensions of trust,
including perceived reliability, task competence, and affective comfort.
However, they differ in their conceptual emphasis: the TPS-HRI primarily
operationalizes trust as a reflective judgement of system performance
(i.e., ``What percent of the time was the robot reliable''), whereas the
TI-HRC scale emphasizes trust as an experienced, embodied response
arising during interaction (i.e., ``The way the robot moved made me feel
uneasy''). Despite this complementarity, both measures rely on
retrospective self-report and may be insensitive to moment-to-moment
trust dynamics as collaboration unfolds. For this reason, questionnaire
data were interpreted alongside behavioural and interaction-level
measures.

\subsubsection{Objective Measures}\label{objective-measures}

Self-report measures were complemented by objective interaction metrics
derived from system logs and manually coded dialogue transcripts. These
measures captured the behavioural and process-level dynamics of
collaboration, including interaction timing, task performance, speech
recognition quality, and the quality of spoken-language exchange are
aspects of the interaction that self-report alone would not fully
characterise.

Coded metrics were derived from interaction logs and manually coded
dialogue transcripts. For the purposes of dialogue coding, a turn was
defined as a participant utterance and the robot's subsequent response,
treated as a single exchange unit. Table~\ref{tbl-post-dialogue-turns}
outlines objective task metrics including the average number of dialogue
turns, session duration, response time (robot dialogue latency), and the
number of engaged versus frustrated responses the robot detected --- all
obtained from log data. Coded metrics from manual transcript analysis
included the proportion of turns characterized by communication
breakdowns --- defined as turns in which the ASR pipeline returned
fragmented or linguistically incoherent output that prevented the robot
from extracting sufficient content to respond meaningfully. This
definition excludes silent turns, in which no speech was detected within
a given timeframe and the robot issued a condition-appropriate check-in;
silence was treated as a distinct interaction event and coded
separately. Other coded metrics included human and robot reasoning,
robot helpful and unhelpful guidance, robot and human affective
engagement, and other task-relevant human and robot contributions. See
Appendix C for a full coding scheme.

\subsection{Participants}\label{sec-participants}

A total of 29 participants were recruited from the Laurentian University
community via word of mouth and the SONA recruitment system. Eligibility
criteria required being 18 years or older, fluent in spoken and written
English, and having normal or corrected-to-normal hearing and vision.
Participants received a \$15 gift card as compensation for their time;
some students additionally received partial credit for participating
depending on their program of study. All procedures were approved by the
Laurentian University Research Ethics Board (ROMEO File: \#6021966).

Although English fluency was an eligibility requirement, in-person
sessions revealed meaningful variability in participants' functional
spoken-language proficiency that was not captured in the online
screening stage. In several sessions, the researcher observed difficulty
in communication with participants prior to the robot interaction;
suggesting that self-reported or screener-assessed fluency did not
always reflect the level of spoken-language proficiency required for
real-time autonomous speech recognition. Given the spoken-language
demands of the task, the researcher documented these observations in
session notes as a methodological record, in anticipation of potential
downstream effects on interaction quality.

Subsequent review of interaction transcripts and system logs confirmed
that a subset of sessions exhibited severe and sustained communication
failure, characterized by fragmented or unintelligible ASR output and
stalled dialogue. In these sessions, the robot was unable to extract
sufficient linguistic content to establish a conversation, respond
meaningfully to participant input, or support task progression. These
interaction failures are described in detail in the Analytic Strategy
and Results sections.

\subsubsection{Randomization Check}\label{randomization-check}

Participants were initially assigned to the responsive or control
condition by Qualtrics using an automated randomization procedure at the
time of sign-up. However, because several scheduled participants did not
attend their in-person session, replacement participants were assigned
to the next available session slot rather than re-randomized. As a
result, the final condition assignment should be considered semi-random
rather than strictly randomized.

To assess whether this deviation introduced systematic bias, we
conducted randomization checks comparing participants across conditions
on key baseline variables. Across analyses, participants in the
responsive and control conditions were comparable with respect to
demographic characteristics, prior experience with robots, and baseline
attitudes toward robots, including gender, age, Negative Attitudes
Toward Robots (NARS), and Need for Cognition scores (see
Table~\ref{tbl-pre}; \textcite{cacioppo1982}). These patterns were
consistent across both the eligible and full analytic samples,
suggesting that the final group composition remained well balanced
despite the partial breakdown of the intended randomization procedure.

\begin{table}

\caption{\label{tbl-pre}Baseline participant characteristics by
interaction policy (full sample)}

\centering{

\fontsize{10.0pt}{12.0pt}\selectfont
\begin{tabular*}{\linewidth}{@{\extracolsep{\fill}}lcccc}
\toprule
\textbf{Characteristic} & \textbf{N} & \textbf{CONTROL}  N = 13\textsuperscript{\textit{1}} & \textbf{RESPONSIVE}  N = 16\textsuperscript{\textit{1}} & \textbf{p-value}\textsuperscript{\textit{2}} \\ 
\midrule\addlinespace[2.5pt]
{\bfseries Gender} & 27 &  &  & 0.84 \\ 
    Woman &  & 6 / 13 (46\%) & 7 / 14 (50\%) &  \\ 
    Man &  & 7 / 13 (54\%) & 7 / 14 (50\%) &  \\ 
{\bfseries Age Group} & 27 &  &  & 0.35 \\ 
    18-24 &  & 5 / 13 (38\%) & 7 / 14 (50\%) &  \\ 
    25-34 &  & 4 / 13 (31\%) & 2 / 14 (14\%) &  \\ 
    34-44 &  & 1 / 13 (7.7\%) & 4 / 14 (29\%) &  \\ 
    45+ &  & 3 / 13 (23\%) & 1 / 14 (7.1\%) &  \\ 
{\bfseries Program} & 25 &  &  & >0.99 \\ 
    Psychology &  & 1 / 13 (7.7\%) & 1 / 12 (8.3\%) &  \\ 
    Engineering &  & 2 / 13 (15\%) & 1 / 12 (8.3\%) &  \\ 
    Computer Science &  & 7 / 13 (54\%) & 6 / 12 (50\%) &  \\ 
    Earth Sciences &  & 0 / 13 (0\%) & 1 / 12 (8.3\%) &  \\ 
    Other &  & 3 / 13 (23\%) & 3 / 12 (25\%) &  \\ 
{\bfseries Experience with Robots} & 29 & 7 / 13 (54\%) & 4 / 16 (25\%) & 0.14 \\ 
{\bfseries Native English Speaker} & 29 &  &  & 0.53 \\ 
    Native English &  & 5 / 13 (38\%) & 8 / 16 (50\%) &  \\ 
    Non-Native English &  & 8 / 13 (62\%) & 8 / 16 (50\%) &  \\ 
{\bfseries NARS Overall} & 29 & 38 (8) & 38 (7) & 0.79 \\ 
{\bfseries Need for Cognition} & 29 & 3.62 (0.78) & 3.74 (0.74) & 0.55 \\ 
{\bfseries Dialogue Viability} & 29 &  &  & 0.63 \\ 
    exclude &  & 3 / 13 (23\%) & 2 / 16 (13\%) &  \\ 
    include &  & 10 / 13 (77\%) & 14 / 16 (88\%) &  \\ 
\bottomrule
\end{tabular*}
\begin{minipage}{\linewidth}
\textsuperscript{\textit{1}}n / N (\%); Mean (SD)\\
\textsuperscript{\textit{2}}Pearson's Chi-squared test; Fisher's exact test; Wilcoxon rank sum test\\
\end{minipage}

}

\end{table}%

\subsection{Analytic Strategy}\label{analytic-strategy}

Analyses were structured to explicitly account for interaction
viability, given the spoken-language demands of the task and the
variability in communication quality observed across sessions (see
Section~\ref{sec-participants}). Sessions were classified as non-viable
when severe communication breakdown prevented sustained dialogue between
the participant and the robot, rendering the experimental manipulation
inoperative.

Communication viability was operationalized using a dialogue-level
metric derived from manual coding of system logs. For each session, the
proportion of dialogue turns affected by speech-recognition failure
(i.e., fragmented or unintelligible utterances) was computed. For the
purposes of dialogue coding, a turn was defined as a participant
utterance and the robot's subsequent response, treated as a single
exchange unit. Sessions in which more than 60\% of dialogue turns were
affected were classified as non-viable, reflecting cases in which
spoken-language interaction could not be meaningfully sustained. This
criterion closely aligned with sessions independently flagged by the
researcher during administration. It should be noted that because
communication viability may itself be influenced by interaction policy,
excluding non-viable sessions may have introduced selection bias and
inflated estimated policy effects.

For this reason, analyses were conducted using three complementary
approaches. Primary analyses were performed on an eligible sample
excluding non-viable sessions, reflecting interactions in which the
spoken-language protocol and experimental manipulation operated as
intended. Full-sample analyses including all sessions were conducted as
sensitivity checks. Finally, mechanism-focused analyses compared viable
and non-viable sessions on interaction-process metrics (e.g., ASR
failure rates, dialogue turn completion, task abandonment) to
characterise how severe communication breakdown alters interaction
dynamics. Trust measures from non-viable sessions were not interpreted
as valid estimates of human-robot trust under functional interaction, as
the robot was unable to sustain dialogue or collaborative behaviour in
these cases.

All analyses were conducted using R (version 4.5.1) within the Quarto
framework. Data manipulation and visualization utilized the tidyverse
suite of packages \autocite{wickham2019}, with mixed-effects models
fitted using the lme4 and lmerTest packages
\autocite{bates2015,kuznetsova2017} while Bayesian hierarchical models
were fitted using the brms package \autocite{burkner2018}. Summary
tables were generated using the gtsummary package
\autocite{sjoberg2021}.

\section{Results}\label{results}

Of the 29 completed sessions, 5 met the pre-specified criterion for
non-viable interaction due to severe and persistent communication
failure. These sessions were characterized by high rates of ASR failure,
incomplete dialogue sequences, and skipped task stages. Primary results
are therefore reported for the eligible sample (n = 24), with
full-sample and mechanism-focused analyses reported as sensitivity and
exploratory analyses, respectively.

\subsection{Descriptive Analysis: Eligible
Sample}\label{descriptive-analysis-eligible-sample}

\begin{figure}

\centering{

\pandocbounded{\includegraphics[keepaspectratio]{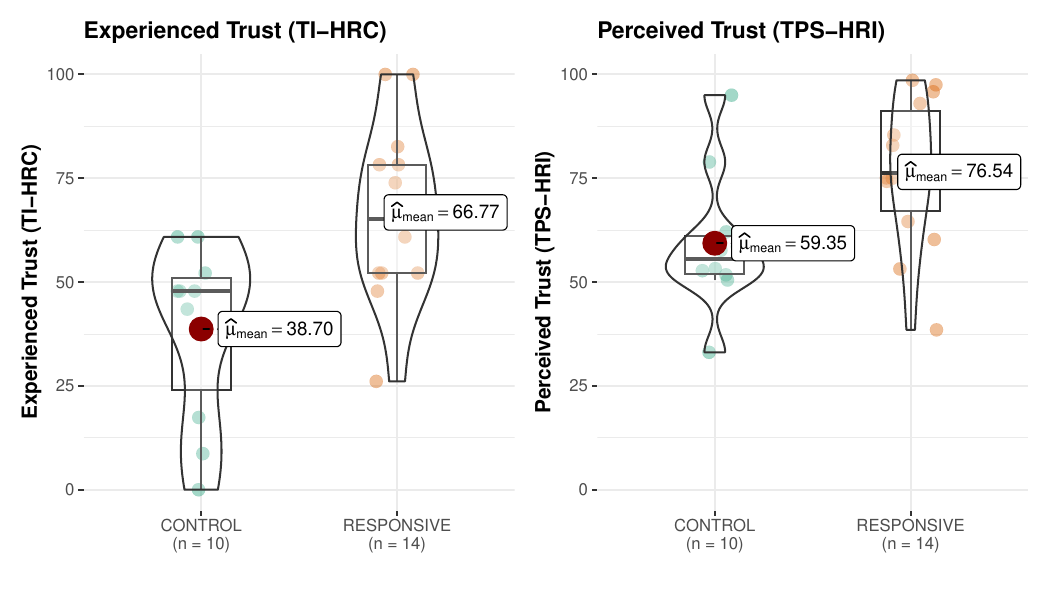}}

}

\caption{\label{fig-post-eligible}Distribution of Trust Perception by
interaction policy. Points represent individual observations; violins
depict score distributions. Red points indicate group means with 95\%
confidence intervals. Statistical comparisons are reported in the
Results section.}

\end{figure}

Simple descriptive comparisons of post-interaction trust measures
indicated higher trust ratings in the responsive condition relative to
the control condition across both trust scales (see
Table~\ref{tbl-post-eligible}). As Figure~\ref{fig-post-eligible} shows,
average raw post-interaction scores on the Trust in Industrial
Human-Robot Collaboration scale (TI-HRC) differed by approximately 26
points (Likert 1-5 converted to 0-100 scale for easier comparison across
scales), while differences on the Trust Perception Scale-HRI (TPS-HRI)
were approximately 15 points higher in the responsive condition compared
to the control. Overall task accuracy did not differ significantly
between conditions (detailed task performance results are reported
below), suggesting that observed differences in trust were not driven by
differential task success but rather by the quality of the interaction
process itself.

\begin{table}

\caption{\label{tbl-post-eligible}Post-Interaction Trust and Task
Accuracy (eligible sample)}

\centering{

\fontsize{10.0pt}{12.0pt}\selectfont
\begin{tabular*}{\linewidth}{@{\extracolsep{\fill}}lccc}
\toprule
\textbf{Characteristic} & \textbf{CONTROL}  N = 10\textsuperscript{\textit{1}} & \textbf{RESPONSIVE}  N = 14\textsuperscript{\textit{1}} & \textbf{p-value}\textsuperscript{\textit{2}} \\ 
\midrule\addlinespace[2.5pt]
Experienced Trust (TI-HRC) & 39 (22) & 67 (21) & {\bfseries 0.004} \\ 
Subscales &  &  &  \\ 
    Reliability & 40 (24) & 65 (18) & {\bfseries 0.012} \\ 
    Trust Perception & 42 (23) & 60 (22) & 0.075 \\ 
    Affective Trust & 50 (31) & 79 (22) & {\bfseries 0.018} \\ 
Perceived Trust (TPS-HRI) & 59 (17) & 77 (18) & {\bfseries 0.022} \\ 
Overall Task Accuracy & 0.60 (0.21) & 0.66 (0.23) & 0.47 \\ 
\bottomrule
\end{tabular*}
\begin{minipage}{\linewidth}
\textsuperscript{\textit{1}}Mean (SD)\\
\textsuperscript{\textit{2}}Wilcoxon rank sum test; Wilcoxon rank sum exact test\\
\end{minipage}

}

\end{table}%

Analysis of dialogue interaction patterns confirmed successful
manipulation of the interaction policies (see
Table~\ref{tbl-post-dialogue-turns}). Manual coding of dialogue
transcripts revealed substantial and significant differences in robot
behaviour across conditions (see Appendix C for coding scheme). In the
responsive condition, the robot employed encouragement in 36\% of
dialogue turns (e.g., ``You're asking great questions!'') compared to
0\% in the control condition (p \textless{} .001), expressed empathy or
acknowledged participant affect in 13\% of turns (e.g., ``I sense you're
frustrated, but we can figure this out together!'') versus 0\% in the
control condition (p \textless{} .001), and used collaborative language
(e.g., ``we,'' ``let's'') in 42\% of turns compared to 5\% in the
control group (p \textless{} .001). While both conditions included
proactive check-ins following periods of participant silence, the
control robot engaged in such check-ins at a higher rate (21\% vs.~13\%
of turns, p = .033), which reflects both the lower levels of
collaborative engagement from participants resulting in more silence and
the control condition's reactive policy that limited assistance to
structured silence monitoring.

Importantly, the proportion of turns coded with communication breakdowns
(i.e., sentence fragments or complete misunderstandings) did not differ
between conditions (25\% vs.~22\%, p = .70), indicating speech
recognition viability among eligible participants was the same between
groups. Participants in the responsive condition also exhibited higher
levels of AI-detected engagement during interaction (i.e., the AI
``perceived'' participants as responsive and interactive in
collaboration), with an average of 3.50 engaged responses (SD = 1.95)
compared to 2.00 (SD = 2.21) in the control condition. Consistent with
these dialogue differences, interactions in the responsive condition
were characterized by longer session durations and slower robot response
times which reflects the latency and time associated with executing
additional dialogue and affective support behaviours. As such,
responsiveness in the present study reflects a bundled interaction
profile rather than a single isolated manipulation. Observed trust
differences may therefore partially reflect differences in interaction
exposure or dialogue richness in addition to affect-adaptive behaviour.
Future work should more carefully disentangle these components.

\begin{table}

\caption{\label{tbl-post-dialogue-turns}Post-Interaction Objective
Measure Averages by Policy in the Eligible Sample}

\centering{

\fontsize{10.0pt}{12.0pt}\selectfont
\begin{tabular*}{\linewidth}{@{\extracolsep{\fill}}lccc}
\toprule
\textbf{Characteristic} & \textbf{CONTROL}  N = 10\textsuperscript{\textit{1}} & \textbf{RESPONSIVE}  N = 14\textsuperscript{\textit{1}} & \textbf{p-value}\textsuperscript{\textit{2}} \\ 
\midrule\addlinespace[2.5pt]
Objective Measure Averages &  &  &  \\ 
    Dialogue Turns & 34 (9) & 33 (5) & 0.45 \\ 
    Session Duration (min) & 13.24 (3.06) & 15.26 (2.12) & 0.084 \\ 
    Response Time (ms) & 14.37 (3.76) & 17.24 (2.52) & {\bfseries <0.001} \\ 
    Silent Periods & 5.60 (1.96) & 4.71 (2.05) & 0.29 \\ 
    Engaged Responses & 2.00 (2.21) & 3.50 (1.95) & {\bfseries 0.040} \\ 
    Frustrated Responses & 0.60 (0.70) & 0.93 (1.21) & 0.68 \\ 
\% of Dialogue Turns Characterized by... &  &  &  \\ 
    Comm. Breakdowns & 0.25 (0.17) & 0.22 (0.16) & 0.70 \\ 
Human Affective Engagement & 0.04 (0.05) & 0.10 (0.08) & {\bfseries 0.032} \\ 
    Human Reasoning & 0.26 (0.12) & 0.35 (0.16) & 0.21 \\ 
    Robot Reasoning & 0.12 (0.07) & 0.37 (0.14) & {\bfseries <0.001} \\ 
    Robot Helpful Guidance & 0.68 (0.08) & 0.84 (0.08) & {\bfseries <0.001} \\ 
    Robot Unhelpful Contributions & 0.08 (0.04) & 0.02 (0.03) & {\bfseries 0.003} \\ 
    Robot Encouragement & 0.00 (0.00) & 0.36 (0.11) & {\bfseries <0.001} \\ 
    Robot Empathy Expression & 0.00 (0.00) & 0.13 (0.09) & {\bfseries <0.001} \\ 
    Robot Collaborative Language & 0.05 (0.05) & 0.42 (0.16) & {\bfseries <0.001} \\ 
    Robot Proactive Check-ins & 0.21 (0.08) & 0.13 (0.08) & {\bfseries 0.033} \\ 
\bottomrule
\end{tabular*}
\begin{minipage}{\linewidth}
\textsuperscript{\textit{1}}Mean (SD)\\
\textsuperscript{\textit{2}}Wilcoxon rank sum test; Wilcoxon rank sum exact test\\
\end{minipage}

}

\end{table}%

Beyond descriptive comparisons, hierarchical models were fitted to
evaluate the robustness of interaction policy effects while controlling
for baseline covariates and accounting for measurement structure.

\subsection{Hierarchical Models}\label{hierarchical-models}

Given the pilot nature of the study and the modest sample size,
hierarchical modeling was used to estimate effect direction and
associated uncertainty rather than to support definitive confirmatory
claims. A Bayesian framework was adopted to allow partial pooling across
trust items and sessions, improving parameter stability under limited
data. Results are therefore interpreted in terms of effect magnitude and
posterior uncertainty rather than binary significance thresholds.

To operationalize this approach, mixed-effects models were fitted using
both frequentist and Bayesian estimation frameworks. All models included
interaction policy (responsive vs.~control) as the primary fixed effect,
with baseline negative attitudes toward robots (NARS) and native English
fluency included as covariates. Random intercepts for sessions and trust
items accounted for the repeated-measures structure. Frequentist models
provided hypothesis tests against null effects, whereas Bayesian models
quantified posterior uncertainty and enabled evaluation across varying
levels of communication viability.

Model building proceeded by comparing a baseline model containing
interaction policy alone against models incorporating theoretically
motivated covariates. Adding NARS scores significantly improved model
fit (\(\chi^2\) = 4.82, p = .028), whereas prior experience with robots
did not. Native English fluency did not significantly improve model fit
but was retained due to its relevance for spoken-language interaction
viability.

Bayesian models used weakly informative, scale-appropriate priors. The
intercept was assigned a Normal(50, 25) prior, reflecting the 0-100
trust score range and centering estimates near the midpoint of the
scale. Fixed effects were assigned Normal(0, 10) priors to constrain
effects to plausible magnitudes. Random-effect and residual standard
deviations were assigned Exponential(1) priors to regularize variance
estimates while retaining flexibility. All models demonstrated
satisfactory convergence (\(\hat{R}\) \(\leq\) 1.01; effective sample
sizes \textgreater{} 1000).

\subsubsection{Primary Analysis: Eligible Sample (n =
24)}\label{primary-analysis-eligible-sample-n-24}

Frequentist linear mixed-effects models revealed consistent positive
effects of responsive interaction on both trust measures. For
experienced trust (TI-HRC), participants who interacted with the
responsive robot reported significantly higher post-interaction trust
than those in the control condition (\(\beta\) = 16.28, SE = 5.14, t =
3.17, p = .005). Higher baseline negative attitudes toward robots were
associated with lower trust scores (\(\beta\) = -7.43, SE = 2.81, p =
.016), while native English fluency was not significantly associated
with trust. Inclusion of random intercepts for individual trust items
improved model fit, indicating meaningful item-level variability beyond
session-level differences. For perceived trust (TPS-HRI), a comparable
pattern emerged, with responsive interaction associated with higher
trust scores (\(\beta\) = 14.17, SE = 6.50, t = 2.00, p = .046).
However, random intercepts for trust items did not improve model fit for
this scale, likely reflecting differences in scale format and response
interface: TPS-HRI was administered using a continuous slider input via
touchpad, whereas TI-HRC employed discrete Likert-style response
options. The slider-based interface may have reduced response precision
and item-level variability, though meaningful between-condition
differences remained detectable at the aggregate level.

\begin{figure}

\pandocbounded{\includegraphics[keepaspectratio]{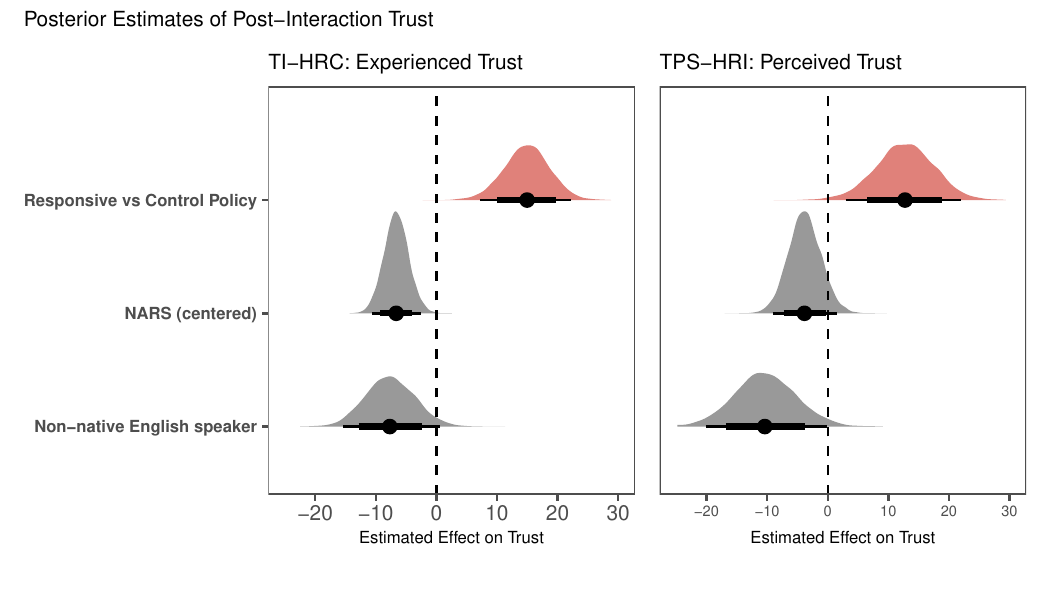}}

\caption{\label{fig-finalmodels-posterior}Posterior distributions of
fixed effects from the final Bayesian mixed-effects models (eligible
sample: n=24) predicting post-interaction trust. Fixed effects include
policy, Negative Attitudes Towards Robots (NARS) and language (binary).
Half-eye densities show posterior distributions; points indicate
posterior medians; thick and thin intervals denote 80\% and 95\%
credible intervals, respectively. The dashed vertical line marks a null
effect, and positive values indicate higher trust scores on the
respective scale (points on a 0-100 scale).}

\end{figure}

Bayesian estimation converged on similar effect magnitudes with high
posterior certainty. As illustrated in
Figure~\ref{fig-finalmodels-posterior}, for TI-HRC, the responsive
condition showed a posterior median effect of \(\beta\) = 14.98 (95\%
credible interval {[}7.29, 22.22{]}), with near-unity probability of a
positive effect. For TPS-HRI, the posterior median was \(\beta\) = 12.76
(95\% credible interval {[}2.96, 22.06{]}), with posterior probability
exceeding 99\% that the effect was positive. As expected baseline NARS
showed credible negative associations with both outcomes, while native
English fluency showed negative associations that were credible for
TPS-HRI but uncertain for TI-HRC. Model fit was substantial for TPS-HRI
(conditional \(R^2\) = 0.64) and moderate for TI-HRC (conditional
\(R^2\) = 0.42), with fixed effects explaining 16\% and 21\% of
variance, respectively. The smaller item-level variance for TI-HRC
suggests greater coherence among affective trust items under functional
interaction conditions.

\subsubsection{Sensitivity Analysis: Full Sample (n =
29)}\label{sensitivity-analysis-full-sample-n-29}

To assess robustness, Bayesian models were refitted including all
sessions regardless of communication viability. The responsive
interaction effect remained positive for both trust measures but showed
substantial attenuation compared to the eligible sample. For TPS-HRI,
the posterior median effect was \(\beta\) = 7.04 (95\% credible interval
{[}-1.83, 15.67{]}). Although uncertainty increased and the credible
interval included zero, the posterior probability of a positive effect
remained high (\textgreater94\%). Model fit decreased relative to the
eligible sample (conditional \(R^2\) = 0.44), indicating increased
unexplained variability when sessions with severe communication
breakdown were included. For TI-HRC, attenuation was more pronounced.
The posterior median effect decreased to \(\beta\) = 7.17 (95\% credible
interval {[}-1.97, 16.70{]}), with reduced probability of a large
effect. Model fit remained moderate (conditional \(R^2\) = 0.60), but
residual variance increased, consistent with the inclusion of
interactions in which collaborative behaviour could not be sustained.
These results indicate that experienced trust is particularly sensitive
to interaction breakdown, and that trust ratings obtained under
non-functional interaction conditions do not reflect graded variation in
collaborative experience but rather reflect the collapse of the
interaction itself.

\subsubsection{Mechanism Analysis: Communication Breakdown
(n=29)}\label{mechanism-analysis-communication-breakdown-n29}

\begin{figure}

\pandocbounded{\includegraphics[keepaspectratio]{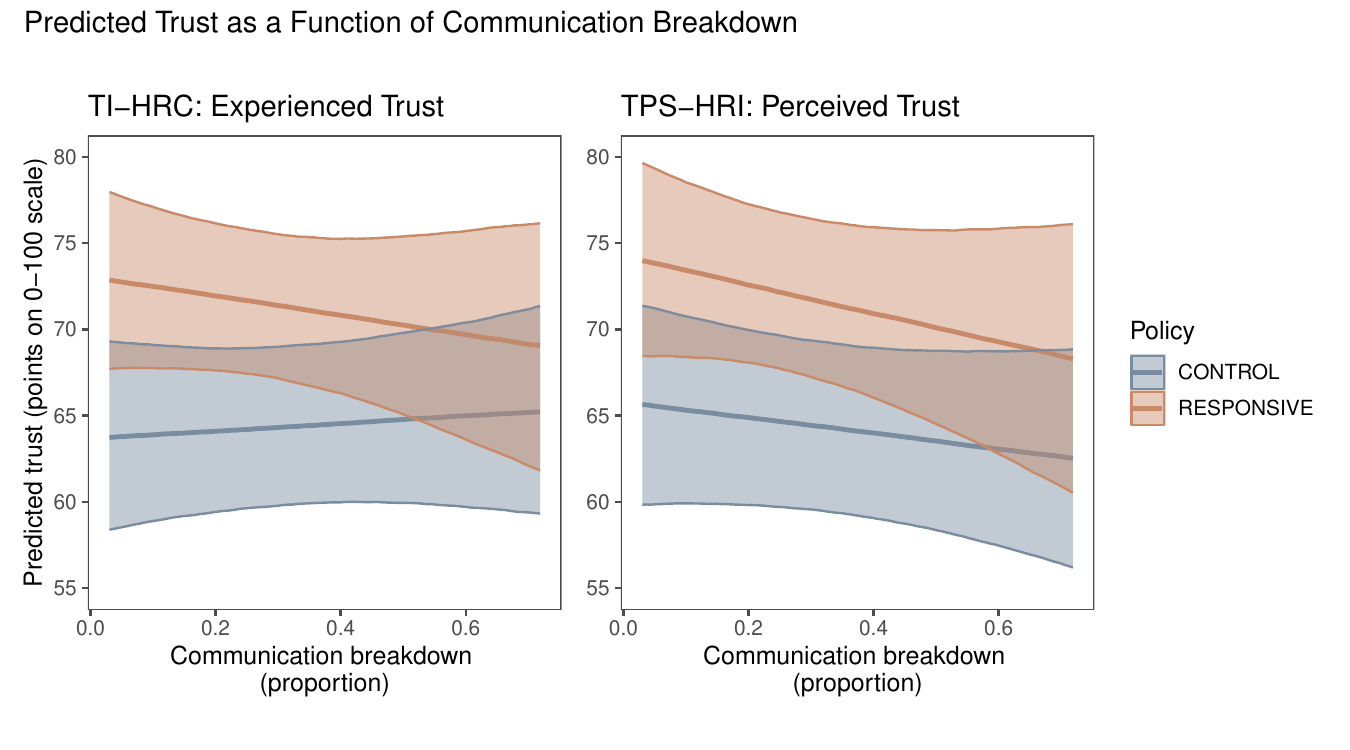}}

\caption{\label{fig-mech-posterior}Predicted trust as a function of
communication breakdown by interaction policy. Panels show experienced
trust (left) and perceived trust (right). Lines represent posterior
medians and shaded bands indicate 80\% credible intervals. Communication
breakdown is defined as the proportion of interaction turns flagged as
breakdowns; predictions are shown at mean baseline negative attitudes
toward robots.}

\end{figure}

To examine whether communication quality moderated the effect of
interaction policy, Bayesian models were fitted in the full sample with
the proportion of turns characterized by communication breakdown
included as an interaction term. Predicted trust as a function of
breakdown proportion for each condition and trust measure is shown in
Figure~\ref{fig-mech-posterior}.

The two trust dimensions responded differently to communication
degradation. For experienced trust (TI-HRC), the responsive condition
showed consistently higher trust than control at low breakdown levels,
but this advantage attenuated as breakdown increased (posterior median
\(\beta\) = -5.97, 95\% CI {[}-23.01, 10.92{]}; 76\% probability
negative). This pattern is consistent with the interpretation that
experienced trust depends on the robot's sustained ability to deliver
responsive behaviour --- when communication fails frequently, the
affect-adaptive features that define the responsive policy cannot be
expressed, and the trust advantage erodes. In contrast, perceived trust
(TPS-HRI) showed a weak, unstable interaction centered near zero
(posterior median \(\beta\) = -1.14, 95\% CI {[}-18.87, 16.28{]}; 55\%
probability negative), suggesting evaluative reliability judgements are
less sensitive to graded communication degradation; at least in this
small sample. This may reflect the fact that perceived trust is anchored
by discrete moments of successful collaboration: a robot that continues
attempting to engage despite communication difficulties may still signal
competence, even if sustained rapport cannot be established.

These results clarify the sensitivity analysis findings. The attenuation
of the responsive policy effect in the full sample reflects the
inclusion of sessions in which the communication level required to
deliver the responsive manipulation had substantially failed (defined as
\textgreater60\% of turns affected by ASR failure) and is not simply
noise from increased variability. This supports both the exclusion
criterion applied in the primary analysis and the broader interpretation
that experienced trust, in particular, is contingent on sustained
interaction quality. Put plainly, a robot that cannot understand you
cannot adapt to you and without adaptation, the responsive policy offers
nothing the control policy does not.

\subsection{Task performance}\label{task-performance}

Overall task accuracy did not differ significantly between conditions
(60\% vs.~66\%, p = .534), nor did performance on individual task
components reach statistical significance (see
Table~\ref{tbl-post-tasks}). However, a notable pattern emerged when
examining task structure. For the suspect identification task (Task 1),
which required robot collaboration to complete accurately, participants
in the responsive condition achieved more than double the accuracy of
those in the control condition (64\% vs.~30\%, p = .106), representing a
large effect that approached but did not reach conventional significance
thresholds given the pilot sample size. In contrast, performance on the
location identification task components (building, zone, and floor
identification), where robot assistance was optional, showed no
consistent directional advantage for either condition. These patterns
suggest that responsive robot behaviour may particularly benefit
collaborative performance on tasks requiring sustained interaction and
mutual grounding, though larger samples are needed to establish
statistical reliability. Critically, the absence of significant overall
accuracy differences indicates that observed trust differences cannot be
attributed simply to differential task success, but rather reflect
distinct responses to the interaction process itself.

\begin{table}

\caption{\label{tbl-post-tasks}Post-Interaction Task Performance by
Policy in the Eligible Sample}

\centering{

\fontsize{10.0pt}{12.0pt}\selectfont
\begin{tabular*}{\linewidth}{@{\extracolsep{\fill}}lccc}
\toprule
\textbf{Characteristic} & \textbf{CONTROL}  N = 10\textsuperscript{\textit{1}} & \textbf{RESPONSIVE}  N = 14\textsuperscript{\textit{1}} & \textbf{p-value}\textsuperscript{\textit{2}} \\ 
\midrule\addlinespace[2.5pt]
\% Task Accuracy & 0.60 (0.21) & 0.66 (0.23) & 0.47 \\ 
Suspect ID Accuracy (robot dependent) & 3 / 10 (30\%) & 9 / 14 (64\%) & 0.10 \\ 
Building ID Accuracy & 7 / 10 (70\%) & 11 / 14 (79\%) & 0.67 \\ 
Zone ID Accuracy & 5 / 10 (50\%) & 4 / 14 (29\%) & 0.40 \\ 
Floor ID Accuracy & 7 / 10 (70\%) & 13 / 14 (93\%) & 0.27 \\ 
\bottomrule
\end{tabular*}
\begin{minipage}{\linewidth}
\textsuperscript{\textit{1}}Mean (SD); n / N (\%)\\
\textsuperscript{\textit{2}}Wilcoxon rank sum test; Pearson's Chi-squared test; Fisher's exact test\\
\end{minipage}

}

\end{table}%

\subsection{Individual differences and correlational
patterns}\label{individual-differences-and-correlational-patterns}

The following correlational analyses are exploratory in nature. A broad
set of associations was examined across behavioural, interactional, and
self-report measures, and p-values were not adjusted for multiple
comparisons. These analyses are therefore intended to identify
potentially meaningful patterns and generate hypotheses for future work
rather than to provide confirmatory statistical tests.

\begin{figure}

\centering{

\pandocbounded{\includegraphics[keepaspectratio]{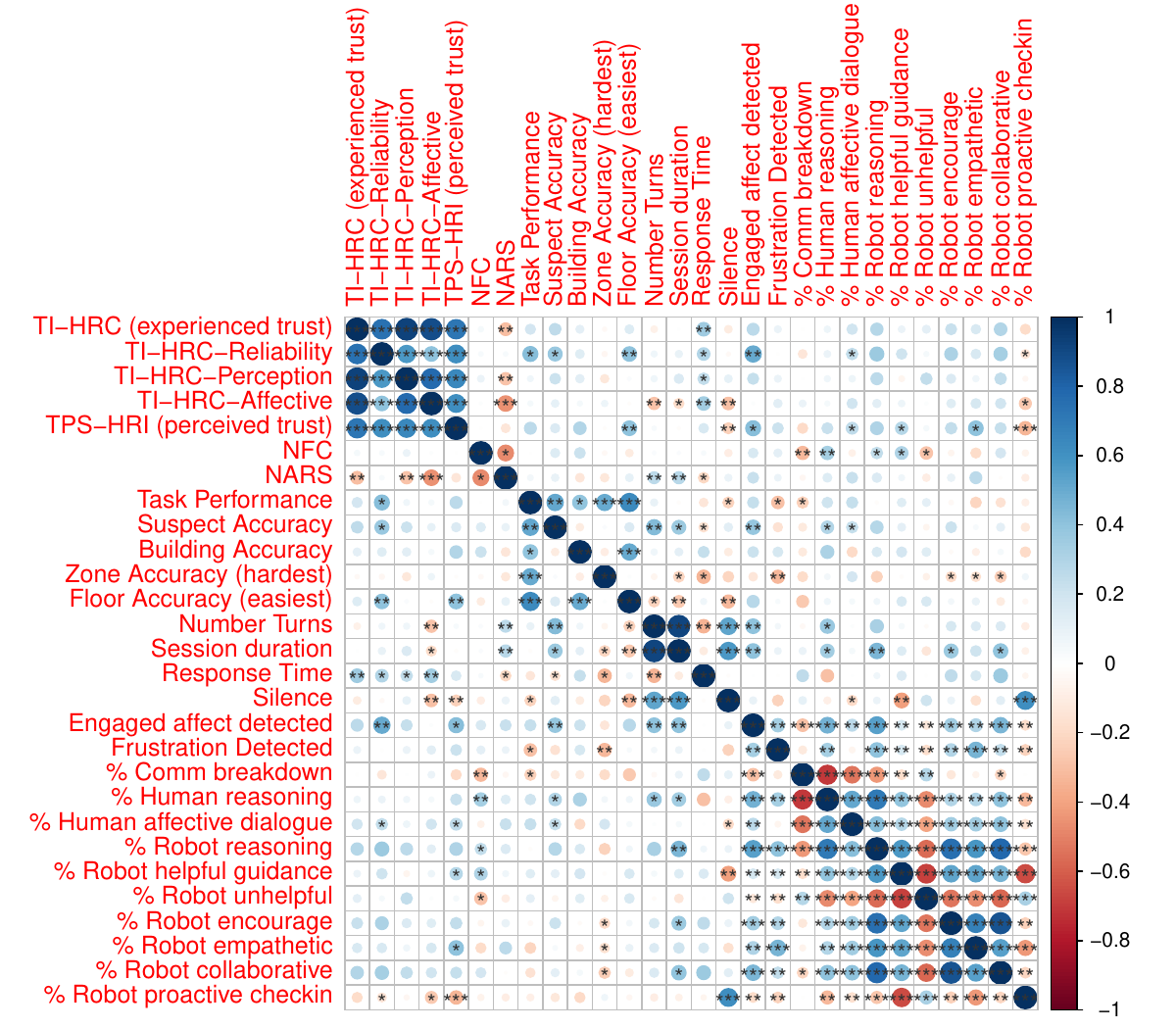}}

}

\caption{\label{fig-corr-matrix}Correlation Matrix of Key Study
Variables}

\end{figure}

Correlational analyses on the full sample (n=29) revealed patterns
consistent with the interpretation that trust was shaped more strongly
by interaction dynamics than by stable individual predispositions or
objective task outcomes. A full correlation matrix is provided in
Figure~\ref{fig-corr-matrix} to visualize the broader pattern of
associations.

As expected, higher Need for Cognition (NFC) scores were negatively
associated with baseline Negative Attitudes Towards Robots (NARS; r =
-.48, p = .01), indicating that individuals who enjoy effortful thinking
tend to hold more positive prior attitudes toward robots. However,
neither NFC nor NARS showed significant associations with
post-interaction trust outcomes (r = -.17 to -.33, p \textgreater{}
.11), suggesting that the impact of responsive robot behaviour was
relatively independent of participants' baseline dispositions toward
robots or cognitive engagement preferences. At the same time, higher NFC
scores were strongly associated with greater engagement in reasoning
dialogue during collaboration (r = .57, p = .003), indicating that
cognitive engagement tendencies may have influenced interaction
behaviour even if they did not directly determine trust evaluations (see
Figure~\ref{fig-corr-matrix}).

Consistent with the experimental manipulation, specific robot dialogue
behaviours showed substantial correlations with trust outcomes. Robot
empathy expression---quantified as the proportion of dialogue turns in
which the robot acknowledged participant affect or expressed
understanding (e.g., ``I sense your frustrated, how can I help?'')---was
strongly correlated with both experienced trust (TI-HRC; r = .53, p =
.008) and perceived trust (TPS-HRI: r = .65, p = .001). Similarly, robot
use of collaborative language (e.g., ``we,'' ``let's'') was positively
associated with experienced trust (TI-HRC: r = .50, p = .012), as was
robot encouragement (TI-HRC: r = .45, p = .027; TPS-HRI: r = .42, p =
.040). These associations suggest that the specific affective and
collaborative behaviours implemented in the responsive condition were
linked to participants' trust evaluations beyond simple condition
assignment.

Participant engagement during interaction, operationalized as the
frequency of detected positive affective responses, was positively
correlated with perceived trust (r = .52, p = .009) and showed a
trending association with experienced trust (r = .30, p = .156). Engaged
responses (both AI detected and affective dialogue by the human) were
associated with longer interaction duration (r = .54, p = .007), greater
use of collaborative language by the robot (r = .50, p = .013), and
fewer communication breakdowns (r = -.49, p = .016). This pattern
suggests that responsive robot behaviour may have fostered a reciprocal
dynamic in which robot affective adaptation elicited participant
engagement, particularly in those high in NFC, which in turn supported
smoother interaction and higher trust.

Notably, objective task performance showed no significant association
with either full trust measure (suspect accuracy: r = .18-.20, p
\textgreater{} .35; overall accuracy: r = -.05 to .19, p \textgreater{}
.37), but did show a significant negative association with the TI-HRC
Reliability subscale (r = .40, p \textless{} .05), as well as a negative
association with detected frustration, silent periods and communication
breakdowns. The dissociation between task outcomes and overall trust
ratings suggests that participants' trust judgments primarily reflected
the quality and affective tone of the interaction process rather than
instrumental task success. Task performance was not included as a
covariate in primary models to avoid conditioning on a potential
mediator of interaction policy effects.

\section{Discussion}\label{discussion}

This pilot study examined how a responsive interaction policy influences
trust during embodied, fully autonomous, spoken-language human-robot
collaboration. Unlike much prior work in HRI trust research, all
dialogue management, affect inference, task progression, and response
generation were executed autonomously and in real time, without
Wizard-of-Oz control or scripted repair. Participants were therefore
exposed not only to adaptive interaction behaviour, but also to the
latency, recognition errors, and coordination failures characteristic of
deployed autonomous systems. The study was conducted as a pilot, with
the primary aim of assessing feasibility and clarifying mechanisms under
these realistic constraints rather than providing definitive
confirmatory tests.

Across analytic approaches, the responsive interaction policy was
associated with higher post-interaction trust compared to a neutral,
reactive control policy when interaction remained viable. These
differences emerged without reliable differences in overall task
accuracy. Although the robot-dependent task showed a descriptive
advantage for the responsive condition, this effect did not reach
statistical significance in the present pilot sample. Taken together,
this pattern suggests that trust judgments were shaped more strongly by
the quality and tone of the interaction process than by instrumental
task success alone. This is consistent with theoretical accounts that
conceptualize trust as an emergent property of interaction dynamics
rather than a static attribute of the system or a simple function of
performance outcomes \autocite{lee2004,devisser2020}.

Trust in human-robot interaction is widely recognized as
multidimensional, encompassing perceptions of reliability and competence
as well as affective comfort and perceived collaboration
\autocite{lee2004,devisser2020}. The use of two validated trust measures
in the present study allowed these components to be examined in
parallel. Rather than treating trust as a static attribute of the
system, the results suggest that it functioned as an evaluation of the
collaborative experience as it unfolded over time. Notably, affective
and experiential components of trust were more sensitive to interaction
degradation than reliability-focused evaluations, reinforcing the view
that trust in autonomous HRI is shaped by process-level interaction
dynamics rather than performance outcomes alone.

These differences became particularly apparent under conditions of
reduced communication viability. When speech recognition failures
accumulated and conversational grounding could not be re-established,
the trust advantage associated with responsiveness diminished. In these
degraded sessions, the responsive policy continued generating proactive
assistance and repair-oriented behaviour. However, when linguistic
grounding could not be restored, these behaviours did not reliably
recover collaboration and may instead have increased cognitive load
relative to the neutral policy's more restrained style. This pattern
suggests that affect-responsive policies operate within a viability
window: they enhance trust when grounding is intact but cannot
compensate for sustained communicative breakdown.

A key aspect of this study is that these dynamics were observed under
full autonomy. Because failures were not covertly repaired by a human
experimenter, communication viability became an explicit feature of the
interaction rather than a hidden nuisance variable. This visibility
clarifies an important constraint: trust formation in spoken-language
HRI presupposes sufficient linguistic grounding. When that grounding
collapses, trust ratings may no longer reflect calibrated evaluations of
competence or benevolence, but instead the breakdown of the
collaborative process itself.

Taken together, the findings support a process-oriented account of trust
in autonomous HRI. Trust appears to emerge from how a system manages
uncertainty, repair, and coordination over time, rather than from task
performance alone. This pilot study identifies both promising effects of
affect-responsive interaction policies under viable conditions and clear
constraints imposed by communication breakdown. These insights provide a
concrete foundation for the design of subsequent studies aimed at
stabilizing language grounding and more precisely isolating the
mechanisms through which responsiveness shapes collaborative experience.

\section{Limitations}\label{limitations}

Several limitations constrain interpretation of the present findings and
point directly to areas for refinement.

The study was conducted as a pilot with a small sample size, limiting
statistical power for detecting higher-order interactions involving task
structure, communication quality, and individual differences. Although
effect sizes were often substantial and broadly consistent across
analytic approaches, uncertainty remains high. Replication with larger
samples will be necessary to establish the stability of these effects
and to more robustly test proposed mediation pathways.

Spoken-language interaction viability emerged as a central constraint,
revealing an important direction for future research. Although English
fluency was an eligibility requirement, substantial variability in
functional spoken-language proficiency was observed during in-person
sessions. In a subset of interactions, persistent speech recognition
failure prevented the experimental manipulation from operating as
intended, leading to exclusion from primary analyses on methodological
grounds. Rather than reflecting a shortcoming of the experimental
design, these cases highlight a fundamental challenge for autonomous
spoken-language HRI: trust and collaboration presuppose a minimum level
of linguistic grounding, and when that grounding fails, higher-level
interaction constructs are no longer meaningfully instantiated.

The visibility of this constraint is itself a consequence of autonomy.
In scripted or Wizard-of-Oz paradigms, language breakdown can be
covertly repaired or masked by human intervention. In fully autonomous
systems, communication viability becomes an explicit property of the
interaction that must be detected, managed, and responded to by the
robot. This points to a clear research direction focused on interaction
policies that recognize emerging language mismatch and adapt
accordingly, rather than assuming linguistic competence as a fixed
prerequisite.

Natural language understanding was also constrained by the task-specific
policy design used in the robot-dependent task. Fixed mappings between
participant questions and predefined task features meant that
semantically valid but unexpected phrasing, synonym use, or
multi-attribute queries occasionally led to misinterpretation or
incorrect responses (e.g., treating ``orange hair'' as distinct from the
ground-truth feature ``red hair''). These failures reflect limitations
in prompt design and natural language understanding and robustness
rather than participant reasoning and likely contributed to some
interaction breakdowns.

Measurement-related factors may have introduced additional noise. The
TPS-HRI trust instrument (perceived trust) relied on a continuous
slider-based input administered via a laptop touchpad, whereas the
TI-HRC relied on a discrete, clickable, Likert-style response.
Touchpad-based slider interaction can be awkward and imprecise for some
users, which may have attenuated effects on the continuous scale
relative to the Likert-based measure. Future work should consider
standardizing response modalities across trust measures to reduce
measurement noise and improve sensitivity as well as incorporating
additional process-level measures of trust dynamics during interaction
(e.g., real-time trust ratings, physiological indicators) to complement
post-interaction evaluations.

An additional limitation concerns the absence of a non-embodied
comparison condition. The present study examined responsiveness within
an embodied robotic platform, but did not include a functionally
equivalent conversational agent presented through a simple chat
interface. As a result, it remains unclear to what extent the observed
trust effects are attributable to the interaction policy itself, the
physical embodiment of the robot, or the interaction between embodiment
and responsiveness. Future work should include a minimal conversational
control condition in order to disentangle policy-level effects from
embodiment-driven influences and to determine whether similar
process-level trust dynamics emerge in purely virtual interaction
contexts.

Finally, affect inference in the deployed system relied primarily on
speech-based signals and conversational context. This design choice
prioritized real-time stability and robustness under autonomous
constraints but necessarily limited the richness of affect sensing.
Incorporating facial expression or prosodic features could improve
responsiveness, though such approaches introduce additional latency,
orchestration complexity, and failure modes that were beyond the scope
of this pilot study.

\section{Conclusions and Future Work}\label{conclusions-and-future-work}

This pilot study provides evidence that affect-responsive interaction
policies can influence trust during fully autonomous, in-person
human-robot collaboration. The observed effects emerged independently of
overall task accuracy, reinforcing the central role of interaction
quality and affective responsiveness in shaping collaborative
experience.

At the same time, the findings delineate clear boundary conditions.
Trust advantages were most pronounced under viable interaction
conditions and attenuated when communication breakdown accumulated. When
communicative grounding collapsed, differences between interaction
policies diminished and trust ratings became less interpretable as
calibrated evaluations. Communication viability thus appears to function
as a prerequisite for trust formation in spoken-language HRI rather than
merely as a moderating variable.

These results directly inform the next phase of work at both empirical
and systems levels. Larger samples will enable more precise estimation
of effects and formal testing of mediation pathways linking
responsiveness, interaction fluency, affective engagement, and
differentiated trust components. Designs that better stabilize language
handling will allow responsiveness to be evaluated under more controlled
grounding conditions.

Architecturally, future iterations will move toward more explicit
state-based or graph-oriented dialogue management, allowing conditional
stage transitions, structured repair strategies, and flexible tool
integration. Adaptive language management mechanisms will be explored to
detect emerging communication mismatch, such as repeated recognition
failures or repair loops, and to adjust linguistic complexity, pacing,
or response format accordingly. These refinements are intended to expand
the range of conditions under which affect-responsive policies can
operate effectively.

More broadly, this pilot study underscores the importance of evaluating
trust in autonomous systems under conditions that expose, rather than
conceal, system limitations. Understanding how trust is shaped,
constrained, and potentially restored under real autonomy conditions is
essential for the responsible deployment of social robots in
collaborative environments.

\section{Ethical Approval}\label{ethical-approval}

All procedures were approved by the Laurentian University Research
Ethics Board (ROMEO File: \#6021966). Participants provided written
informed consent prior to participation and were debriefed in-person
following the session. Data were de-identified and stored securely in
accordance with institutional guidelines.

\section{Data and Code Availability}\label{data-and-code-availability}

The analysis code (R) is available at:
\href{https://github.com/shaunaheron2/misty-paper}{GitHub Repository}.
The interaction system and prompt templates used to implement the
experimental conditions (Python) are also available in the same
repository. De-identified participant data (survey scales and
session-level summary measures) are available upon reasonable request to
the corresponding author. Raw interaction logs are not publicly
available due to participant privacy constraints.

\section{Conflict of Interest}\label{conflict-of-interest}

The authors declare no conflicts of interest.

\section{Author Contributions}\label{author-contributions}

S. Heron: Conceptualization, Methodology, Software, Formal Analysis,
Investigation, Data Curation, Writing - Original Draft, Writing - Review
\& Editing, Visualization.

M.C. Lau: Conceptualization, Methodology, Writing - Review \& Editing,
Supervision.

\section{Funding}\label{funding}

This work was supported by IAMGOLD President's Innovation Fund (Grant
No.~2025-20377).

\section{Declaration of Generative AI and AI-assisted Technologies in
the Writing
Process}\label{declaration-of-generative-ai-and-ai-assisted-technologies-in-the-writing-process}

During the preparation of this work, the author(s) used Claude (Sonnet
4.6, Anthropic) to review the manuscript for clarity and readability.
After reviewing the AI-generated suggestions, the author(s) made
revisions to the text as appropriate. The author(s) takes full
responsibility for the content of the published work.

\section{References}\label{refs}

\printbibliography[heading=none]

@inproceedings{lin2022,
	title = {2022 31st IEEE International Conference on Robot and Human Interactive Communication (RO-MAN)},
	author = {Lin, Ting-Han and Ng, Spencer and Sebo, Sarah},
	year = {2022},
	month = {08},
	date = {2022-08},
	pages = {37--44},
	doi = {10.1109/RO-MAN53752.2022.9900828},
	url = {https://ieeexplore.ieee.org/document/9900828},
	note = {ISSN: 1944-9437}
}

@article{maure,
	title = {Autonomy in socially assistive robotics: a systematic review},
	author = {Maure, Romain and Bruno, Barbara},
	journal = {Frontiers in Robotics and AI},
	year = {2025},
	pages = {1586473},
	volume = {12},
	doi = {10.3389/frobt.2025.1586473},
	url = {https://pmc.ncbi.nlm.nih.gov/articles/PMC12491022/},
	note = {PMID: 41050743
PMCID: PMC12491022}
}

@inbook{schaefer2016,
	title = {Measuring Trust in Human Robot Interactions: Development of the {\textquotedblleft}Trust Perception Scale-HRI{\textquotedblright}},
	author = {Schaefer, Kristin E.},
	editor = {Mittu, Ranjeev and Sofge, Donald and Wagner, Alan and Lawless, W.F.},
	year = {2016},
	date = {2016},
	publisher = {Springer US},
	pages = {191--218},
	url = {https://doi.org/10.1007/978-1-4899-7668-0_10},
	note = {DOI: 10.1007/978-1-4899-7668-0{\_}10},
	address = {Boston, MA},
	langid = {en}
}

@article{charalambous2016,
	title = {The Development of a Scale to Evaluate Trust in Industrial Human-robot Collaboration},
	author = {Charalambous, George and Fletcher, Sarah and Webb, Philip},
	year = {2016},
	month = {04},
	date = {2016-04-01},
	journal = {International Journal of Social Robotics},
	pages = {193--209},
	volume = {8},
	number = {2},
	doi = {10.1007/s12369-015-0333-8},
	url = {https://doi.org/10.1007/s12369-015-0333-8},
	langid = {en}
}

@article{emaminejad2022,
	title = {Trustworthy AI and robotics: Implications for the AEC industry},
	author = {Emaminejad, Newsha and Akhavian, Reza},
	year = {2022},
	month = {07},
	date = {2022-07-01},
	journal = {Automation in Construction},
	pages = {104298},
	volume = {139},
	doi = {10.1016/j.autcon.2022.104298},
	url = {https://www.sciencedirect.com/science/article/pii/S0926580522001716}
}

@article{campagna2025,
	title = {A Systematic Review of Trust Assessments in Human{\textendash}Robot Interaction},
	author = {Campagna, Giulio and Rehm, Matthias},
	year = {2025},
	month = {01},
	date = {2025-01-27},
	journal = {J. Hum.-Robot Interact.},
	pages = {30:1{\textendash}30:35},
	volume = {14},
	number = {2},
	doi = {10.1145/3706123},
	url = {https://doi.org/10.1145/3706123}
}

@article{devisser2020,
	title = {Towards a Theory of Longitudinal Trust Calibration in Human{\textendash}Robot Teams},
	author = {de Visser, Ewart J. and Peeters, Marieke M. M. and Jung, Malte F. and Kohn, Spencer and Shaw, Tyler H. and Pak, Richard and Neerincx, Mark A.},
	year = {2020},
	month = {05},
	date = {2020-05-01},
	journal = {International Journal of Social Robotics},
	pages = {459--478},
	volume = {12},
	number = {2},
	doi = {10.1007/s12369-019-00596-x},
	url = {https://doi.org/10.1007/s12369-019-00596-x},
	langid = {en}
}

@article{nicolas2021,
	title = {The personality of anthropomorphism: How the need for cognition and the need for closure define attitudes and anthropomorphic attributions toward robots},
	author = {Nicolas, Spatola and Agnieszka, Wykowska},
	year = {2021},
	month = {09},
	date = {2021-09-01},
	journal = {Computers in Human Behavior},
	pages = {106841},
	volume = {122},
	doi = {10.1016/j.chb.2021.106841},
	url = {https://www.sciencedirect.com/science/article/pii/S0747563221001643}
}

@article{fu2021,
	title = {Design and Research of Intelligent Safety Monitoring Robot for Coal Mine Shaft Construction},
	author = {Fu, Wenjun and Xu, Ying and Liu, Liangping and Zhang, Liang},
	editor = {Tsai, Sang-Bing},
	year = {2021},
	month = {01},
	date = {2021-01},
	journal = {Advances in Civil Engineering},
	pages = {6897767},
	volume = {2021},
	number = {1},
	doi = {10.1155/2021/6897767},
	url = {https://onlinelibrary.wiley.com/doi/10.1155/2021/6897767},
	langid = {en}
}

@inproceedings{shayganfar2019,
	title = {2019 IEEE International Conference on Humanized Computing and Communication (HCC)},
	author = {Shayganfar, Mahni and Rich, Charles and Sidner, Candace and {Hylák}, Benjamin},
	year = {2019},
	month = {09},
	date = {2019-09},
	pages = {7--15},
	doi = {10.1109/HCC46620.2019.00010},
	url = {https://ieeexplore.ieee.org/document/8940829}
}

@article{fartook2025,
	title = {Enhancing emotional support in human-robot interaction: Implementing emotion regulation mechanisms in a personal drone},
	author = {Fartook, Ori and McKendrick, Zachary and Oron-Gilad, Tal and Cauchard, Jessica R.},
	year = {2025},
	month = {05},
	date = {2025-05-01},
	journal = {Computers in Human Behavior: Artificial Humans},
	pages = {100146},
	volume = {4},
	doi = {10.1016/j.chbah.2025.100146},
	url = {https://www.sciencedirect.com/science/article/pii/S2949882125000301}
}

@article{diab2025,
	title = {TICK: A Knowledge Processing Infrastructure for Cognitive Trust in Human{\textendash}Robot Interaction},
	author = {Diab, Mohammed and Demiris, Yiannis},
	year = {2025},
	month = {01},
	date = {2025-01-04},
	journal = {International Journal of Social Robotics},
	pages = {1--33},
	doi = {10.1007/s12369-024-01206-1},
	url = {https://link.springer.com/article/10.1007/s12369-024-01206-1},
	note = {Company: Springer
Distributor: Springer
Institution: Springer
Label: Springer
Publisher: Springer Netherlands},
	langid = {en}
}

@article{cacioppo1982,
	title = {The need for cognition},
	author = {Cacioppo, John T. and Petty, Richard E.},
	year = {1982},
	date = {1982},
	journal = {Journal of Personality and Social Psychology},
	pages = {116--131},
	volume = {42},
	number = {1},
	doi = {10.1037/0022-3514.42.1.116},
	note = {Place: US
Publisher: American Psychological Association}
}

@article{birnbaum2016,
	title = {What robots can teach us about intimacy: The reassuring effects of robot responsiveness to human disclosure},
	author = {Birnbaum, Gurit E. and Mizrahi, Moran and Hoffman, Guy and Reis, Harry T. and Finkel, Eli J. and Sass, Omri},
	year = {2016},
	month = {10},
	date = {2016-10-01},
	journal = {Computers in Human Behavior},
	pages = {416--423},
	volume = {63},
	doi = {10.1016/j.chb.2016.05.064},
	url = {https://www.sciencedirect.com/science/article/pii/S0747563216303910}
}

@article{ciuffreda2025,
	title = {Design and Development of a Technological Platform Based on a Sensorized Social Robot for Supporting Older Adults and Caregivers: GUARDIAN Ecosystem},
	author = {Ciuffreda, Ilaria and Amabili, Giulio and Casaccia, Sara and Benadduci, Marco and Margaritini, Arianna and Maranesi, Elvira and Marconi, Fabrizio and De Masi, Alexander and Alberts, Janna and de Koning, Judith and Cuijpers, Raymond and Revel, Gian Marco and Nap, Henk Herman and Vastenburg, Martijn and Naveira, Alexandra Villaverde and Bevilacqua, Roberta},
	year = {2025},
	month = {05},
	date = {2025-05-01},
	journal = {International Journal of Social Robotics},
	pages = {803--822},
	volume = {17},
	number = {5},
	doi = {10.1007/s12369-023-01038-5},
	url = {https://doi.org/10.1007/s12369-023-01038-5},
	langid = {en}
}

@inproceedings{spitale2023,
	title = {Robotic Mental Well-being Coaches for the Workplace: An In-the-Wild Study on Form},
	author = {Spitale, Micol and Axelsson, Minja and Gunes, Hatice},
	year = {2023},
	month = {03},
	date = {2023-03-13},
	publisher = {Association for Computing Machinery},
	pages = {301{\textendash}310},
	series = {HRI '23},
	doi = {10.1145/3568162.3577003},
	url = {https://dl.acm.org/doi/10.1145/3568162.3577003},
	address = {New York, NY, USA}
}

@misc{mistyrobotics,
  title={Misty-II Robot Platform},
  author={{Furhat Robotics}},
  year={2023},
  url={https://www.mistyrobotics.com},
  note={Accessed 2025-01-XX}
}

@inproceedings{wischnewski2023,
	title = {Measuring and Understanding Trust Calibrations for Automated Systems: A Survey of the State-Of-The-Art and Future Directions},
	author = {Wischnewski, Magdalena and {Krämer}, Nicole and {Müller}, Emmanuel},
	year = {2023},
	month = {04},
	date = {2023-04-19},
	publisher = {Association for Computing Machinery},
	pages = {1{\textendash}16},
	series = {CHI '23},
	doi = {10.1145/3544548.3581197},
	url = {https://doi.org/10.1145/3544548.3581197},
	address = {New York, NY, USA}
}

@article{muir1994,
	title = {Trust in automation: Part I. Theoretical issues in the study of trust and human intervention in automated systems},
	author = {MUIR, BONNIE M.},
	year = {1994},
	month = {11},
	date = {1994-11-01},
	journal = {Ergonomics},
	pages = {1905--1922},
	volume = {37},
	number = {11},
	doi = {10.1080/00140139408964957},
	url = {https://doi.org/10.1080/00140139408964957},
	note = {Publisher: Taylor \& Francis}
}

@article{hancock2011,
	title = {A meta-analysis of factors affecting trust in human-robot interaction},
	author = {Hancock, Peter A. and Billings, Deborah R. and Schaefer, Kristin E. and Chen, Jessie Y. C. and de Visser, Ewart J. and Parasuraman, Raja},
	year = {2011},
	month = {10},
	date = {2011-10},
	journal = {Human Factors},
	pages = {517--527},
	volume = {53},
	number = {5},
	doi = {10.1177/0018720811417254},
	note = {PMID: 22046724},
	langid = {eng}
}

@article{lee2004,
	title = {Trust in automation: designing for appropriate reliance},
	author = {Lee, John D. and See, Katrina A.},
	year = {2004},
	date = {2004},
	journal = {Human Factors},
	pages = {50--80},
	volume = {46},
	number = {1},
	doi = {10.1518/hfes.46.1.50_30392},
	note = {PMID: 15151155},
	langid = {eng}
}

@article{bates2015,
	title = {Fitting Linear Mixed-Effects Models Using lme4},
	author = {Bates, Douglas and {Mächler}, Martin and Bolker, Ben and Walker, Steve},
	year = {2015},
	month = {10},
	date = {2015-10-07},
	journal = {Journal of Statistical Software},
	pages = {1--48},
	volume = {67},
	doi = {10.18637/jss.v067.i01},
	url = {https://doi.org/10.18637/jss.v067.i01},
	langid = {en}
}

@Article{kuznetsova2017,
	title = {{lmerTest} Package: Tests in Linear Mixed Effects Models},
	author = {Alexandra Kuznetsova and Per B. Brockhoff and Rune H. B. Christensen},
	journal = {Journal of Statistical Software},
	year = {2017},
	volume = {82},
	number = {13},
	pages = {1--26},
	doi = {10.18637/jss.v082.i13}
}

@article{wickham2019,
	title = {Welcome to the Tidyverse},
	author = {Wickham, Hadley and Averick, Mara and Bryan, Jennifer and Chang, Winston and McGowan, {Lucy D'Agostino} and {François}, Romain and Grolemund, Garrett and Hayes, Alex and Henry, Lionel and Hester, Jim and Kuhn, Max and Pedersen, Thomas Lin and Miller, Evan and Bache, Stephan Milton and {Müller}, Kirill and Ooms, Jeroen and Robinson, David and Seidel, Dana Paige and Spinu, Vitalie and Takahashi, Kohske and Vaughan, Davis and Wilke, Claus and Woo, Kara and Yutani, Hiroaki},
	year = {2019},
	month = {11},
	date = {2019-11-21},
	journal = {Journal of Open Source Software},
	pages = {1686},
	volume = {4},
	number = {43},
	doi = {10.21105/joss.01686},
	url = {https://joss.theoj.org/papers/10.21105/joss.01686},
	langid = {en}
}

@article{sjoberg2021,
	title = {Reproducible Summary Tables with the gtsummary Package},
	author = {Sjoberg, Daniel D. and Whiting, Karissa and Curry, Michael and Lavery, Jessica A. and Larmarange, Joseph},
	year = {2021},
	month = {06},
	date = {2021-06-22},
	journal = {The R Journal},
	pages = {570--580},
	volume = {13},
	number = {1},
	doi = {10.32614/RJ-2021-053},
	url = {https://doi.org/10.32614/RJ-2021-053/}
}

@Article{burkner2018,
title = {Advanced {Bayesian} Multilevel Modeling with the {R}
	Package {brms}},
	author = {Paul-Christian Bürkner},
	journal = {The R Journal},
	year = {2018},
	volume = {10},
	number = {1},
	pages = {395--411},
	doi = {10.32614/RJ-2018-017},
	encoding = {UTF-8},
  }

@article{choi2025,
	title = {Exploring Challenges and Opportunities in Manufacturing and Intelligence for Future Robotics},
	author = {Choi, Jun Young and Ahn, Semin and Kim, Dohyeon and Heo, Jun and Yun, Won-jae and Hong, Sungjin and Bae, Sunghoon and Ahn, Sung-Hoon},
	year = {2025},
	month = {09},
	date = {2025-09-01},
	journal = {International Journal of Precision Engineering and Manufacturing},
	pages = {2203--2222},
	volume = {26},
	number = {9},
	doi = {10.1007/s12541-025-01318-2},
	url = {https://doi.org/10.1007/s12541-025-01318-2},
	langid = {en}
}

@misc{Chase_LangChain_2022,
  author = {Chase, Harrison},
  title = {{LangChain}},
  url = {https://langchain.com/},
  year = {2022}
  }

@article{zhang2019,
	title = {Theory of Robot Mind: False Belief Attribution to Social Robots in Children With and Without Autism},
	author = {Zhang, Yaoxin and Song, Wenxu and Tan, Zhenlin and Wang, Yuyin and Lam, Cheuk Man and Hoi, Sio Pan and Xiong, Qianhan and Chen, Jiajia and Yi, Li},
	year = {2019},
	month = {08},
	date = {2019-08-09},
	journal = {Frontiers in Psychology},
	volume = {10},
	doi = {10.3389/fpsyg.2019.01732},
	url = {https://www.frontiersin.org/journals/psychology/articles/10.3389/fpsyg.2019.01732/full},
	note = {Publisher: Frontiers},
	langid = {English}
}

@inproceedings{atone2022,
	title = {International Conference on Image Processing and Capsule Networks},
	author = {Atone, Swati A. and Bhalchandra, A. S.},
	year = {2022},
	date = {2022},
	publisher = {Springer, Cham},
	pages = {498--522},
	doi = {10.1007/978-3-030-84760-9_43},
	url = {https://link.springer.com/chapter/10.1007/978-3-030-84760-9_43},
	note = {ISSN: 2367-3389},
	langid = {en}
}

@article{arkin2003,
	title = {An ethological and emotional basis for human{\textendash}robot interaction},
	author = {Arkin, Ronald C. and Fujita, Masahiro and Takagi, Tsuyoshi and Hasegawa, Rika},
	year = {2003},
	month = {03},
	date = {2003-03},
	journal = {Robotics and Autonomous Systems},
	pages = {191--201},
	volume = {42},
	number = {3-4},
	doi = {10.1016/S0921-8890(02)00375-5},
	url = {https://linkinghub.elsevier.com/retrieve/pii/S0921889002003755},
	langid = {en}
}

@misc{nomura,
	title = {Negative Attitudes toward Robots Scale},
	author = {Nomura, Tatsuya and Suzuki, Tomohiro and Kanda, Takayuki and Kato, Kensuke},
	doi = {10.1037/t57930-000},
	langid = {en}
}

@article{wei,
	title = {Chain-of-Thought Prompting Elicits Reasoning in Large Language Models},
	author = {Wei, Jason and Wang, Xuezhi and Schuurmans, Dale and Bosma, Maarten and Ichter, Brian and Xia, Fei and Chi, Ed and Le, Quoc and Zhou, Denny},
	doi = {10.48550/arXiv.2201.11903}
}

@article{mcduff,
	title = {Do Facial Expressions Predict Ad Sharing? A Large-Scale Observational Study},
	author = {McDuff, Daniel and Berger, Jonah},
	doi = {10.48550/arXiv.1912.10311}
}

@Manual{duckdb2026,
  title = {duckdb: DBI Package for the DuckDB Database Management System},
  author = {Hannes Mühleisen and Mark Raasveldt},
  year = {2026},
  note = {R package version 1.4.3.9001},
  url = {https://r.duckdb.org/},
}

@misc{python,
  author       = {{Python Software Foundation}},
  title        = {Python Language Reference, Version 3.10},
  year         = {2023},
  howpublished = {\url{https://www.python.org}},
}

@misc{deepgram,
  author       = {{Deepgram}},
  title        = {Deepgram Speech Recognition API},
  year         = {2024},
  howpublished = {\url{https://deepgram.com}},
}

@misc{gemini,
  author       = {{Google}},
  title        = {Gemini API},
  year         = {2024},
  howpublished = {\url{https://ai.google.dev}},
}

@misc{misty,
	title = {Misty},
	url = {https://docs.mistyrobotics.com/}
}

@article{kok2020,
	title = {Trust in Robots: Challenges and Opportunities},
	author = {Kok, Bing Cai and Soh, Harold},
	year = {2020},
	month = {12},
	date = {2020-12-01},
	journal = {Current Robotics Reports},
	pages = {297--309},
	volume = {1},
	number = {4},
	doi = {10.1007/s43154-020-00029-y},
	url = {https://doi.org/10.1007/s43154-020-00029-y},
	langid = {en}
}

@article{bettencourt2025,
	title = {Investigating the Feasibility of a Wizard-of-Oz Robotic Interface (R2C3) in a Social Skills Group for Children with Autism Spectrum Disorder},
	author = {Bettencourt, Carlotta and Grossard, Charline and Zou, Jianling and Segretain, Marie and Bree, Morgane and Pellerin, Hugues and Anzalone, Salvatore M. and Chetouani, Mohamed and Cohen, David},
	year = {2025},
	month = {07},
	date = {2025-07-01},
	journal = {International Journal of Social Robotics},
	pages = {1395--1411},
	volume = {17},
	number = {7},
	doi = {10.1007/s12369-025-01243-4},
	url = {https://doi.org/10.1007/s12369-025-01243-4},
	langid = {en}
}

@article{val-calvo2020,
	title = {Affective Robot Story-Telling Human-Robot Interaction: Exploratory Real-Time Emotion Estimation Analysis Using Facial Expressions and Physiological Signals},
	author = {Val-Calvo, Mikel and Alvarez-Sanchez, Jose Ramon and Ferrandez-Vicente, Jose Manuel and Fernandez, Eduardo},
	year = {2020},
	date = {2020},
	journal = {IEEE Access},
	pages = {134051--134066},
	volume = {8},
	doi = {10.1109/ACCESS.2020.3007109},
	url = {https://ieeexplore.ieee.org/document/9133100/},
	langid = {en}
}

\section{Appendix A}\label{sec-appendix-a}

\subsection{Technical Implementation
Details}\label{technical-implementation-details}

The experimental system comprised a fully autonomous, multi-stage
collaborative task in which participants interacted with the Misty II
social robot to solve a two-part investigative scenario. Interaction was
mediated through spoken dialogue and a companion web interface, allowing
the robot and participant to jointly reason about task information. The
system was designed as a mixed-initiative dialogue architecture with
optional affect-responsive behaviour, implemented without human
intervention during experimental sessions.

\subsubsection{Hardware Platform}\label{hardware-platform}

The robot platform used in this study was the Misty II social robot.
Misty II is a mobile social robot equipped with an expressive display,
articulated head and arms, and programmable RGB LEDs. These components
were used to produce synchronized verbal and nonverbal behaviour,
including eye expressions, head movements, arm gestures, and
colour-based state indicators. Audio input was captured via the robot's
RTSP video stream, which provided real-time access to the microphone
signal for downstream speech processing.

\subsubsection{Software Architecture}\label{software-architecture}

All system components were implemented in Python (version 3.10)
\autocite{python}. The software architecture integrated robot control,
speech processing, dialogue management, task logic, and data logging
into a single autonomous pipeline. Core dependencies included the Misty
Robotics Python SDK for robot control, the Deepgram SDK for speech
recognition \autocite{misty,deepgram}, FFmpeg for audio stream
processing, Flask and Flask-SocketIO for the web-based task interface,
and DuckDB for structured data logging \autocite{duckdb2026}.

\subsubsection{Dialogue Management and Large Language Model
Integration}\label{dialogue-management-and-large-language-model-integration}

Dialogue was managed using the LangChain framework, which provided
abstraction over message handling, memory persistence, and large
language model integration \autocite{Chase_LangChain_2022}. The system
used Google's Gemini API as the underlying language model, configured to
produce strictly JSON-formatted outputs to ensure reliable downstream
parsing and execution on the robot \autocite{gemini}.

The deployed model was \texttt{gemini-2.5-flash-lite}, selected for its
low-latency response characteristics. Generation temperature was set to
0.7 to balance coherence and variability. Conversation history was
maintained using a buffer-based memory mechanism, allowing the robot to
reference prior exchanges within a session while resetting memory
between participants. Conversation histories were stored as
session-specific JSON files to enable post-hoc analysis and recovery.

\subsubsection{Prompt Structure and Context
Injection}\label{prompt-structure-and-context-injection}

System prompts were constructed dynamically at each dialogue turn. Each
prompt consisted of a system message defining task rules, role
constraints, and output format requirements, followed by the accumulated
conversation history and the current participant utterance. In addition
to transcribed speech, structured contextual variables were injected
into the prompt as JSON fields, including the current task stage,
detected emotion labels, timer expiration flags, and task submission
status. This approach allowed the language model to access environmental
state without embedding control information directly into conversational
text.

\subsection{Speech Processing}\label{speech-processing}

Speech-to-text processing was handled by Deepgram's Nova-2 model using
real-time WebSocket streaming \autocite{deepgram}. The system employed
adaptive endpointing and voice activity detection to support
conversational turn-taking. Endpointing thresholds differed across task
stages, with shorter timeouts during dialogue-driven stages and longer
timeouts during log-reading phases.

Text-to-speech output was generated using Misty II's onboard TTS engine,
which produces a synthetic robotic voice. Although external TTS options
(including OpenAI and Deepgram Aura voices) were implemented and tested,
the onboard voice was selected to reduce latency and avoid introducing
human-like vocal qualities that could independently influence trust
perceptions.

\subsection{Emotion Detection and Affective State
Mapping}\label{emotion-detection-and-affective-state-mapping}

Participant affect was inferred from transcribed utterances using a
DistilRoBERTa-based emotion classification model fine-tuned for
English-language emotion detection. The model produced categorical
predictions (e.g., joy, frustration, anxiety, neutral), which were
mapped to higher-level interaction states such as positive engagement,
irritation, or confusion. In the responsive condition, these inferred
states were used to guide dialogue strategy and nonverbal behaviour
selection.

\subsection{Multimodal Behaviour
Generation}\label{multimodal-behaviour-generation}

The robot's nonverbal behaviour was implemented through a library of
custom action scripts combining facial expressions, LED patterns, arm
movements, and head motions. At each dialogue turn, the language model
selected an expression label from a predefined set, which was then
translated into a coordinated multimodal action. In the responsive
condition, additional backchannel behaviours were triggered during
participant speech, including listening cues and emotion-matched
expressions.

LED colours were used to signal system state to participants. A blue LED
indicated active listening, while a purple LED indicated processing or
speaking.

\begin{figure}

\centering{

\includegraphics[width=1\linewidth,height=\textheight,keepaspectratio]{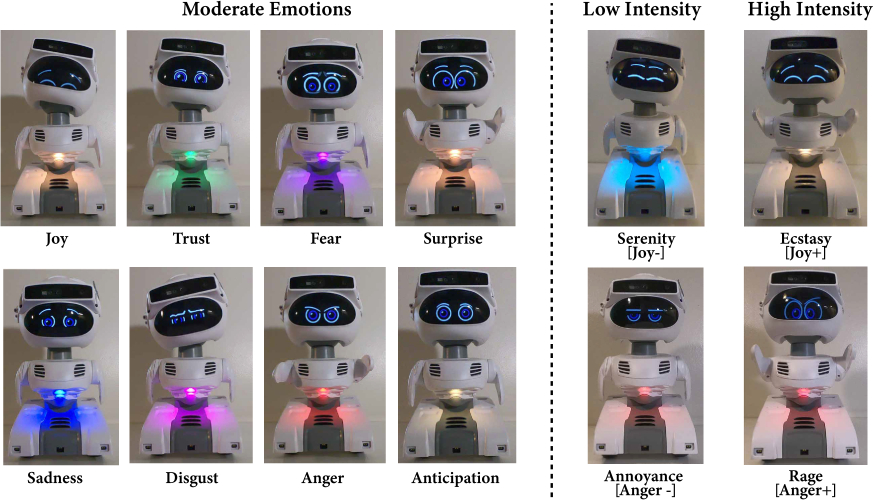}

}

\caption{\label{fig-task1}A selection of facial expressions the language
model could choose from each turn. Retrieved from {[}1{]} N. T. White,
B. Cagiltay, J. E. Michaelis, and B. Mutlu, ``Designing Emotionally
Expressive Social Commentary to Facilitate Child-Robot Interaction,'' in
Interaction Design and Children, Athens Greece: ACM, Jun.~2021,
pp.~314--325. doi: 10.1145/3459990.3460714.}

\end{figure}%

\subsection{Collaborative Tasks}\label{collaborative-tasks}

The interaction consisted of two collaborative tasks inspired by the
puzzle task desigbed by \textcite{lin2022}. In the first task,
participants and the robot jointly solved a ``who-dunnit'' problem by
eliminating suspects from a grid based on yes/no questions. The robot
possessed ground-truth knowledge but was constrained to answering only
feature-based yes/no queries. In the second task, participants and the
robot attempted to locate a missing robot by interpreting cryptic system
and sensor logs. In this task, the robot did not know the solution and
instead provided guidance based on general technical knowledge and
logical reasoning.

Task information and participant responses were presented through a
web-based dashboard. The dashboard displayed suspect grids, system logs,
and response input fields, and communicated task progression events back
to the robot via REST API calls.

\subsection{Data Collection and
Logging}\label{data-collection-and-logging}

All interaction data were logged to a DuckDB relational database
\autocite{duckdb2026}. Logged data included session metadata, turn-level
dialogue transcripts, language model responses, nonverbal behaviour
selections, response latencies, task submissions, detected emotions, and
system events such as stage transitions and timer expirations. This
structure enabled detailed post-hoc analysis of interaction dynamics,
communication failures, and trust-related behaviours.

\subsection{Interaction Dynamics and Control
Policy}\label{interaction-dynamics-and-control-policy}

Two interaction policies were implemented and toggled programmatically
at runtime: a responsive mode and a control mode. In the responsive
mode, the robot proactively offered assistance, adjusted its dialogue
based on inferred affect, and produced supportive backchannel
behaviours. In the control mode, the robot provided general guidance and
more information only when explicitly prompted and did not adapt its
behaviour based on affective cues. The active policy was set prior to
each session and remained fixed throughout the interaction.

Silence handling was implemented using a fixed threshold, after which
the robot issued a check-in prompt. The phrasing of these prompts
differed across conditions to reflect proactive versus reactive
interaction strategies.

\subsection{Inter-process
Communication}\label{inter-process-communication}

System components communicated via a set of Flask-based REST endpoints.
These endpoints synchronized task stage state, detected participant
submissions, managed timer events, and allowed limited facilitator
override when necessary. All communication between the web interface and
the robot occurred locally to ensure low latency and experimental
reliability.

\section{Appendix B}\label{sec-appendix-b}

\subsection{Trust Perception Scale HRI
(TPS-HRI)}\label{trust-perception-scale-hri-tps-hri}

Participants rated the following items on a percentage scale (0-100\%),
indicating the proportion of time each statement applied to the robot
during the interaction.

\begin{itemize}
\tightlist
\item
  What percent of the time was the robot dependable?
\item
  What percent of the time was the robot reliable?
\item
  What percent of the time was the robot responsive?
\item
  What percent of the time was the robot trustworthy?
\item
  What percent of the time was the robot supportive?
\item
  What percent of the time did this robot act consistently?
\item
  What percent of the time did this robot provide feedback?
\item
  What percent of the time did this robot meet the needs of the mission
  task?
\item
  What percent of the time did this robot provide appropriate
  information?
\item
  What percent of the time did this robot communicate appropriately?
\item
  What percent of the time did this robot follow directions?
\item
  What percent of the time did this robot answer the questions asked?
\end{itemize}

\subsection{Trust in Industrial Human-Robot Collaboration
(TI-HRC)}\label{trust-in-industrial-human-robot-collaboration-ti-hrc}

Participants indicated their agreement with the following statements
using a 5-point Likert-type scale (Strongly Disagree to Strongly Agree).
Negatively worded items were reverse-scored prior to analysis.

\textbf{\emph{Reliability}}

\begin{itemize}
\tightlist
\item
  I trusted that the robot would give me accurate answers.
\item
  The robot's responses seemed reliable.
\item
  I felt I could rely on the robot to do what it was supposed to do.
\end{itemize}

\textbf{\emph{Perceptual / Affective Trust}}

\begin{itemize}
\tightlist
\item
  The robot seemed to enjoy helping me.
\item
  The robot was responsive to my needs.
\item
  The robot seemed to care about helping me.
\end{itemize}

\textbf{\emph{Discomfort / Unease}}

\begin{itemize}
\tightlist
\item
  The way the robot moved made me uncomfortable. (R)
\item
  The way the robot spoke made me uncomfortable. (R)
\item
  Talking to the robot made me uneasy. (R)
\end{itemize}

\section{Appendix C}\label{sec-appendix-c}

\subsection{Dialogue Coding Scheme}\label{dialogue-coding-scheme}

\subsubsection{Task Outcome Layer
(Stage-Level)}\label{task-outcome-layer-stage-level}

{\def\LTcaptype{none} 
\begin{longtable}[]{@{}
  >{\raggedright\arraybackslash}p{(\linewidth - 4\tabcolsep) * \real{0.3750}}
  >{\raggedright\arraybackslash}p{(\linewidth - 4\tabcolsep) * \real{0.2500}}
  >{\raggedright\arraybackslash}p{(\linewidth - 4\tabcolsep) * \real{0.3750}}@{}}
\toprule\noalign{}
\begin{minipage}[b]{\linewidth}\raggedright
Variable
\end{minipage} & \begin{minipage}[b]{\linewidth}\raggedright
Type
\end{minipage} & \begin{minipage}[b]{\linewidth}\raggedright
Description
\end{minipage} \\
\midrule\noalign{}
\endhead
\bottomrule\noalign{}
\endlastfoot
\texttt{task\_outcome} & categorical & Final task status
(\texttt{completed}, \texttt{timeout}, \texttt{skipped},
\texttt{partial}, \texttt{abandoned}). \\
\texttt{task\_completed\_without\_help} & binary & Task was completed
without any help requests to the robot. \\
\end{longtable}
}

\subsection{Dialogue Interaction Layer
(Turn-Level)}\label{dialogue-interaction-layer-turn-level}

\subsubsection*{Human Turn Codes}\label{human-turn-codes}
\addcontentsline{toc}{subsubsection}{Human Turn Codes}

{\def\LTcaptype{none} 
\begin{longtable}[]{@{}
  >{\raggedright\arraybackslash}p{(\linewidth - 4\tabcolsep) * \real{0.3750}}
  >{\raggedright\arraybackslash}p{(\linewidth - 4\tabcolsep) * \real{0.2500}}
  >{\raggedright\arraybackslash}p{(\linewidth - 4\tabcolsep) * \real{0.3750}}@{}}
\toprule\noalign{}
\begin{minipage}[b]{\linewidth}\raggedright
Variable
\end{minipage} & \begin{minipage}[b]{\linewidth}\raggedright
Type
\end{minipage} & \begin{minipage}[b]{\linewidth}\raggedright
Description
\end{minipage} \\
\midrule\noalign{}
\endhead
\bottomrule\noalign{}
\endlastfoot
\texttt{human\_help\_request} & binary & Participant explicitly or
implicitly asks the robot for help or guidance. \\
\texttt{human\_reasoning} & binary & Participant reasons out loud with
the robot toward problem-solving. \\
\texttt{human\_confirmation\_seeking} & binary & Participant seeks
confirmation of a tentative belief or solution. \\
\texttt{human\_sentence\_fragment} & binary & Unintelligable sentence
fragment \\
\end{longtable}
}

\subsubsection*{Robot Turn Codes}\label{robot-turn-codes}
\addcontentsline{toc}{subsubsection}{Robot Turn Codes}

{\def\LTcaptype{none} 
\begin{longtable}[]{@{}
  >{\raggedright\arraybackslash}p{(\linewidth - 4\tabcolsep) * \real{0.3889}}
  >{\raggedright\arraybackslash}p{(\linewidth - 4\tabcolsep) * \real{0.2500}}
  >{\raggedright\arraybackslash}p{(\linewidth - 4\tabcolsep) * \real{0.3611}}@{}}
\toprule\noalign{}
\begin{minipage}[b]{\linewidth}\raggedright
Variable
\end{minipage} & \begin{minipage}[b]{\linewidth}\raggedright
Type
\end{minipage} & \begin{minipage}[b]{\linewidth}\raggedright
Description
\end{minipage} \\
\midrule\noalign{}
\endhead
\bottomrule\noalign{}
\endlastfoot
\texttt{robot\_helpful\_guidance} & binary & Robot provides accurate,
task-relevant information or guidance. \\
\texttt{robot\_unhelpful} & binary & Robot provides misleading or
incorrect guidance. \\
\texttt{robot\_stt\_failure} & binary & Robot response reflects a
speech-to-text or input understanding failure. \\
\texttt{robot\_clarification\_request} & binary & Robot asks the
participant for information or to repeat or clarify their input. \\
\texttt{robot\_reasoning} & binary & Robot reasons out loud with the
participant to assist problem-solving \\
\end{longtable}
}

\subsection{Affective Interaction Layer
(Turn-Level)}\label{affective-interaction-layer-turn-level}

\subsubsection*{Robot Affective
behaviour}\label{robot-affective-behaviour}
\addcontentsline{toc}{subsubsection}{Robot Affective behaviour}

{\def\LTcaptype{none} 
\begin{longtable}[]{@{}
  >{\raggedright\arraybackslash}p{(\linewidth - 4\tabcolsep) * \real{0.3889}}
  >{\raggedright\arraybackslash}p{(\linewidth - 4\tabcolsep) * \real{0.2500}}
  >{\raggedright\arraybackslash}p{(\linewidth - 4\tabcolsep) * \real{0.3611}}@{}}
\toprule\noalign{}
\begin{minipage}[b]{\linewidth}\raggedright
Variable
\end{minipage} & \begin{minipage}[b]{\linewidth}\raggedright
Type
\end{minipage} & \begin{minipage}[b]{\linewidth}\raggedright
Description
\end{minipage} \\
\midrule\noalign{}
\endhead
\bottomrule\noalign{}
\endlastfoot
\texttt{robot\_empathy\_expression} & binary & Robot expresses empathy,
encouragement, or reassurance. \\
\texttt{robot\_emotion\_acknowledgement} & binary & Robot explicitly
references an inferred participant emotional state. \\
\texttt{robot\_collaborative\_language} & binary & Robot uses language
like ``we'', ``lets'', ``together''. \\
\end{longtable}
}

\subsubsection*{Human Affective
Response}\label{human-affective-response}
\addcontentsline{toc}{subsubsection}{Human Affective Response}

{\def\LTcaptype{none} 
\begin{longtable}[]{@{}
  >{\raggedright\arraybackslash}p{(\linewidth - 4\tabcolsep) * \real{0.3889}}
  >{\raggedright\arraybackslash}p{(\linewidth - 4\tabcolsep) * \real{0.2500}}
  >{\raggedright\arraybackslash}p{(\linewidth - 4\tabcolsep) * \real{0.3611}}@{}}
\toprule\noalign{}
\begin{minipage}[b]{\linewidth}\raggedright
Variable
\end{minipage} & \begin{minipage}[b]{\linewidth}\raggedright
Type
\end{minipage} & \begin{minipage}[b]{\linewidth}\raggedright
Description
\end{minipage} \\
\midrule\noalign{}
\endhead
\bottomrule\noalign{}
\endlastfoot
\texttt{human\_affective\_engagement} & binary & Participant responds in
a socially warm or engaged manner (e.g., ``It's up to us, Misty!''
and/or mirrors or responds to the robot's affective expression and/or
treats the robot as a social agent. \\
\end{longtable}
}

\end{document}